\documentclass[acmsmall]{acmart}

\AtBeginDocument{%
  \providecommand\BibTeX{{%
    \normalfont B\kern-0.5em{\scshape i\kern-0.25em b}\kern-0.8em\TeX}}}

\setcopyright{acmcopyright}
\copyrightyear{2018}
\acmYear{2018}
\acmDOI{XXXXXXX.XXXXXXX}

\acmJournal{JACM}
\acmVolume{37}
\acmNumber{4}
\acmArticle{111}
\acmMonth{8}

\acmPrice{15.00}
\acmISBN{978-1-4503-XXXX-X/18/06}

\usepackage{graphicx}
\usepackage{subfigure}
\usepackage{multirow}
\usepackage{color}
\usepackage{rotating}
\begin{document}

\title{Cross-BERT for Point Cloud Pretraining}

\author{Xin Li, Peng Li, Zeyong Wei, Zhe Zhu, Mingqiang Wei}
\email{theali942@gmail.com, pengl@nuaa.edu.cn, weizeyong1@gmail.com, zhuzhe0619@nuaa.edu.cn, mingqiang.wei@gmail.com}
\affiliation{%
  \institution{ Nanjing University of Aeronautics and Astronautics}
  \country{China}
}




\author{Junhui Hou}
\email{jh.hou@cityu.edu.hk}
\affiliation{%
  \institution{ City University of Hong Kong}
  \country{Hong Kong SAR}
}

\author{Liangliang Nan}
\email{liangliang.nan@gmail.com}
\affiliation{%
  \institution{ Delft University of Technology}
  \country{Netherlands}
}

\author{Jing Qin}
\email{harry.qin@polyu.edu.hk}
\affiliation{%
  \institution{The Hong Kong Polytechnic University}
  \country{Hong Kong SAR}
}

\author{Haoran Xie}
\email{hrxie@ln.edu.hk}
\affiliation{%
  \institution{Lingnan University}
  \country{Hong Kong SAR}
}

\author{Fu Lee Wang}
\email{pwang@hkmu.edu.hk}
\affiliation{%
  \institution{Hong Kong Metropolitan University}
  \country{Hong Kong SAR}
}

\renewcommand{\shortauthors}{Li, et al.}

\begin{abstract}
Introducing BERT into cross-modal settings raises difficulties in its optimization for handling multiple modalities. Both the BERT architecture and training objective need to be adapted to incorporate and model information from different modalities.
In this paper, we address these challenges by exploring the implicit semantic and geometric correlations between 2D and 3D data of the same objects/scenes. We propose a new cross-modal BERT-style self-supervised learning paradigm, called Cross-BERT. To facilitate pretraining for irregular and sparse point clouds, we design two self-supervised tasks to boost cross-modal interaction. 
The first task, referred to as \textit{Point-Image Alignment}, aims to align features between unimodal and cross-modal representations to capture the correspondences between the 2D and 3D modalities. 
The second task, termed \textit{Masked Cross-modal Modeling}, further improves mask modeling of BERT by incorporating high-dimensional semantic information obtained by cross-modal interaction.
By performing cross-modal interaction, Cross-BERT can smoothly reconstruct the masked tokens during pretraining, leading to notable performance enhancements for downstream tasks. 
Through empirical evaluation, we demonstrate that Cross-BERT outperforms existing state-of-the-art methods in 3D downstream applications. Our work highlights the effectiveness of leveraging cross-modal 2D knowledge to strengthen 3D point cloud representation and the transferable capability of BERT across modalities.
\end{abstract}

\ccsdesc[500]{Computing methodologies~Multimedia information systems}

\keywords{Cross-BERT, point cloud pretraining, geometric processing, self-supervised learning, point-image alignment, masked cross-modal modeling}

\received{20 February 2007}
\received[revised]{12 March 2009}
\received[accepted]{5 June 2009}

\maketitle

\section{Introduction}
\label{sec:intro}
The success of transformers in natural language processing is motivating their applications in the vision community. 
Pretraining with transformers presents challenges due to the limited large-scale labeled data, particularly in the context of 3D vision tasks.
To address it, BERT- and MAE-style methods have emerged as effective solutions.
For example, in the 2D modality, BEiT \cite{bao2021beit}, MAE \cite{he2022masked} and their derivatives \cite{peng2022beit,wei2022masked} can conduct masked image modeling efficiently; in the 3D modality, Point-BERT \cite{DBLP:conf/cvpr/YuTR00L22} reconstructs masked tokens in point clouds directly. In contrast to MAE-style methods, which work to reconstruct each pixel or point, BERT-style methodologies harness the power of discretization more naturally and logically, and assist in mask modeling, with an emphasis on higher-level features. 

However, there is always a performance gap between BERT-style methods and MAE-style methods in both 2D and 3D. We believe there are two causes: (1) Unlike natural language processing (NLP) where a word can be naturally mapped to a token, for random discrete point cloud patches, a single token cannot generalize its semantic features. Traditional point cloud discretization can interfere with the obtained representation; (2) existing BERT-style methods typically focus on processing each modality independently, without exploring the potential benefits of incorporating correlated and complementary information from other modalities. Specifically, 2D rendering provides rich color and texture information but lacks depth and shape details, while 3D point clouds offer precise spatial and geometric information but only sparse and textureless features. If the inherent advantages of the BERT style are to be fully exploited on the irregular and sparse point clouds, it is important to address the problems in discretization and leverage the strengths of 2D-3D correlations. 

\begin{figure}[htbp]
    \begin{center}  
    \includegraphics[width=0.9\linewidth]{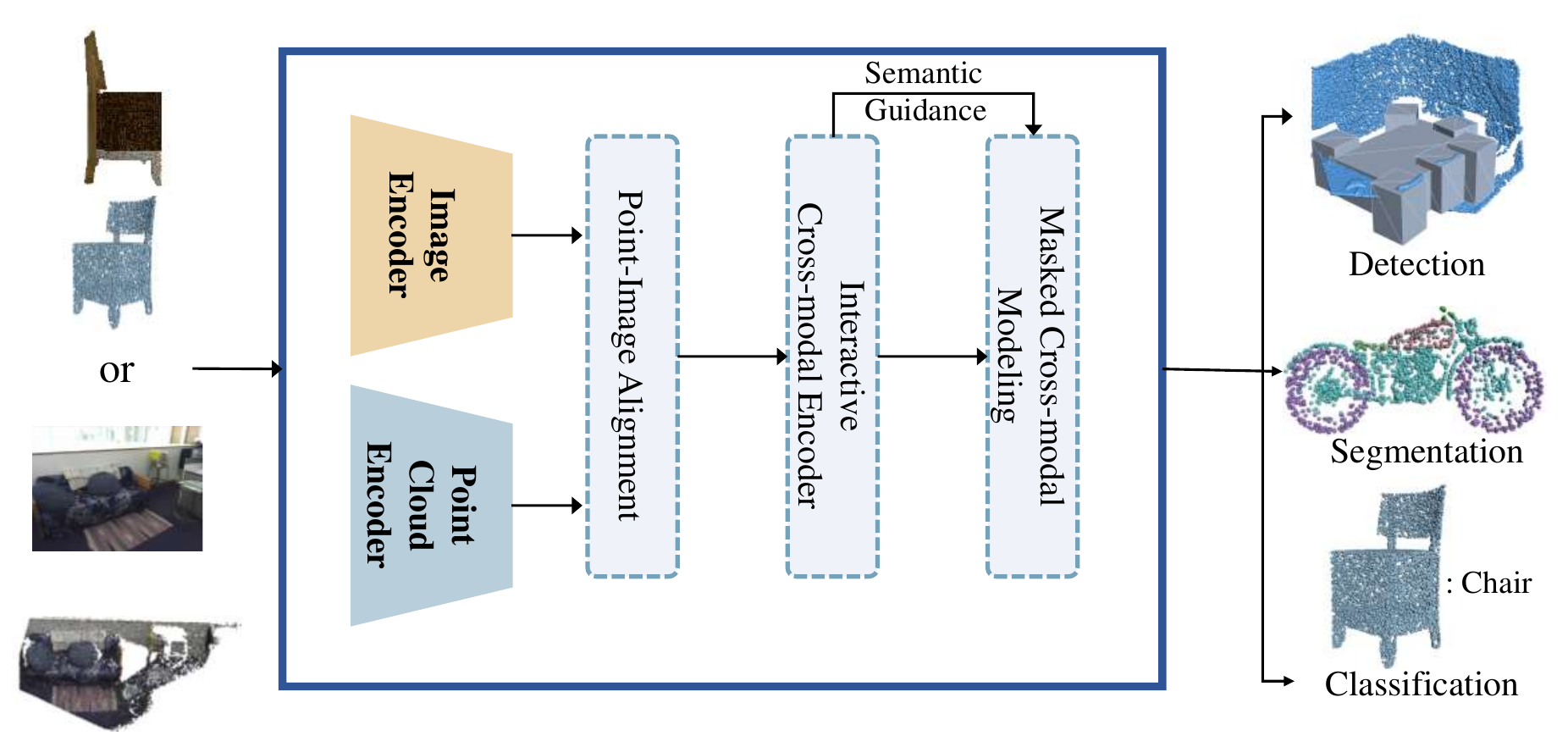}
    \caption{Cross-BERT can learn cross-modal representations via interactive processing and reconstruction of cross-modal data. After diverse pretraining, it excels in various downstream geometric tasks.
    }
    \label{fig:intro}
    \end{center}
\end{figure}

In this paper, we propose Cross-BERT, a cross-modal self-supervised pretraining model designed to incorporate informative content from 2D images into 3D point clouds, leading to high-quality and versatile representations. As shown in Figure \ref{fig:intro},
the approach involves RGB images acquired from different sources to capture 2D expressions and masking 3D tokens with a random ratio to encourage robust feature learning. The two modalities (2D and 3D) are fed into separate unimodal encoders in parallel. To address the challenges of cross-modal integration, we utilise two pretraining tasks: 
(1) \textit{Point-Image Alignment} (PIA) aims to obtain discriminative representations of different modalities for the masked modeling task. The conventional wisdom~\cite{DBLP:conf/cvpr/AfhamDDDTR22} only considers optimization between different modalities, leading to detrimental effects on the integrity of the original unimodal representation.
To this end, we design PIA by integrating both intra-modality and inter-modality contrastive objectives to obtain aligned yet informative features.
(2) \textit{Masked Cross-modal Modeling} (MCM) aims to improve the conventional single-choice mask modeling task with the assistance of cross-modal semantic information.
We utilize the multi-choice masked modeling scheme~\cite{li2022mc} and guide the tokenizer with high-level and non-single semantic information to unravel that the commonly used tokenizer tends to focus on the fine-grained local and single information of each patch, leading to incorrect token prediction. 

\begin{figure*}[t!]
    \begin{center}
    \includegraphics[width=1\linewidth]{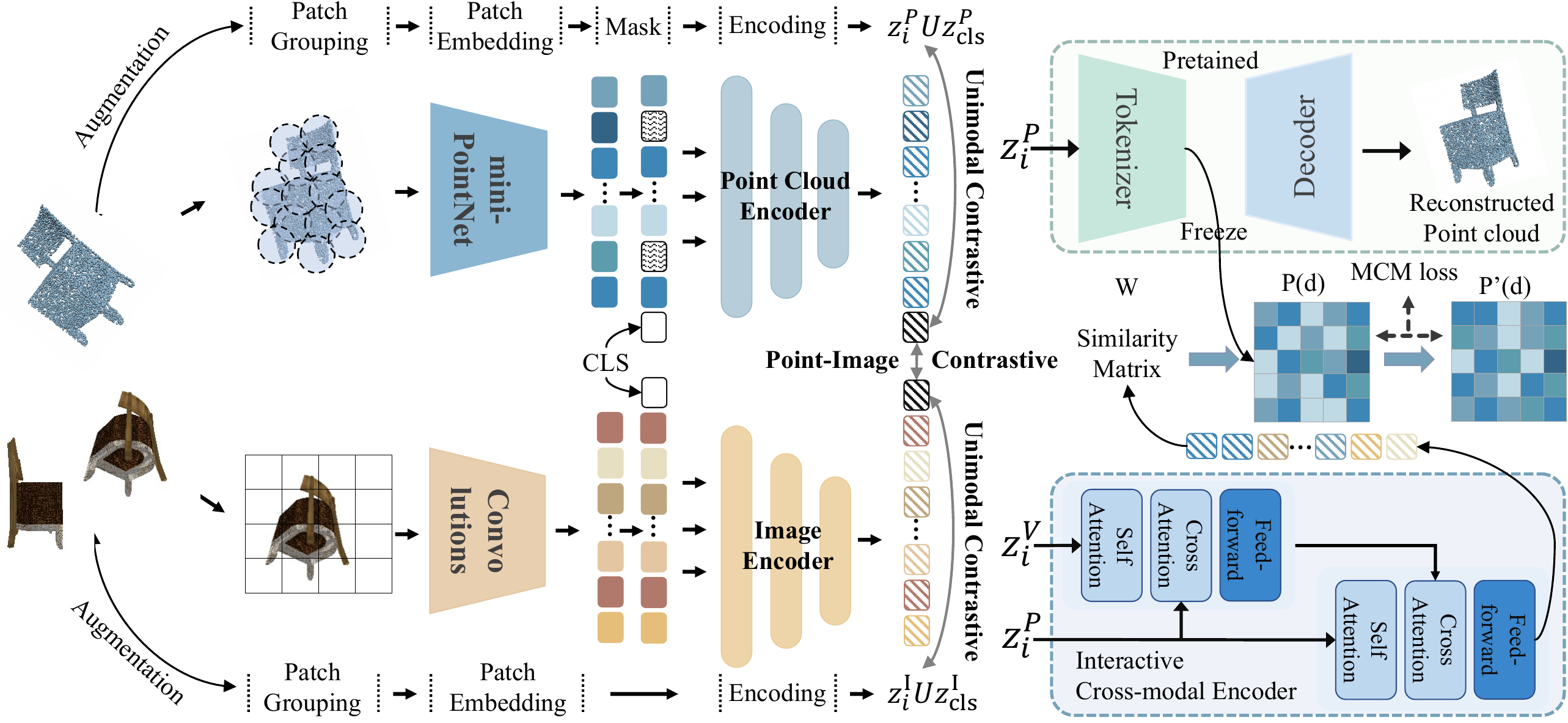}
    \caption{Architecture of Cross-BERT. 3D point clouds and rendered RGB images serve as inputs, being unimodal encoders (Point Cloud Encoder and Image Encoder) for sequential tokens.  We set two pretraining tasks (Point-Image Alignment and Masked Cross-modal Modeling) and three loss functions (PIC, UMC, and MCM) to learn robust and discriminative features.
    }
    \label{fig:framework}
    \end{center}
\end{figure*}
Extensive experiments demonstrate that pretraining on both 2D images and 3D point clouds using Cross-BERT produces high-quality and general representations and achieves state-of-the-art performance on various downstream 3D tasks. 
In summary, our contributions are three-fold: 
\begin{itemize}
\item We propose Cross-BERT, a cross-modal framework for self-supervised pretraining of 3D point clouds with two specially customized tasks.
 
\item We design the novel \textit{Point-Image Alignment} task equipped with multiple contrastive objectives to enhance cross-modal interaction, thereby facilitating the masked modeling task. 

\item To improve unimodal mask modeling, we introduce the \textit{Masked Cross-modal Modeling} task, which leverages high-level semantic information from the \textit{Interactive Cross-modal Encoder}.

\end{itemize}
\section{{Related Works}}

This section reviews recent advances in self-supervised pretraining for point clouds, including generative, contrastive, and cross-modal methods.

\subsection{{Generative Methods}}
Generative methods learn features through self-reconstruction, where a point cloud is first encoded into a feature or distribution and then decoded back to itself \cite{DBLP:conf/cvpr/YangFST18,YonghengZhao20183DPC,DBLP:conf/iccv/HanWLZ19,he2022masked,DBLP:conf/iccv/HassaniH19}. 
Inspiring from MAE~\cite{he2022masked}, Point-MAE \cite{pang2022masked} and Point-M2AE~\cite{zhang2022point} randomly mask patches of input point clouds and reconstruct the missing points. Besides, \cite{10314552} proposes a geometrically and adaptively masked auto-encoder on point clouds, tries to capture more general feature. 
Point-BERT \cite{DBLP:conf/cvpr/YuTR00L22} performs discretization on input point clouds and translates the prediction of 3D coordinates into token IDs.
Besides, an alternative to generative methods is to use generative adversarial networks for generative modeling \cite{han2019view, achlioptas2018learning}.

\subsection{{Contrastive Methods}}
For contrastive methods, they learn point cloud representations by maximizing the similarity between positive pairs of augmented point clouds and minimizing the similarity between negative pairs. 
PointContrast \cite{DBLP:conf/cvpr/XieLCT18} proposes a pretext task in which the representations of a single point cloud from different views are encouraged to remain consistent, focusing on high-level scene understanding tasks. 
Some methods attempt to combine contrastive learning and other self-supervised strategies to obtain more robust representations. For example, ClusterNet~\cite{DBLP:conf/3dim/ZhangZ19} combines contrastive learning and offline clustering for effective representation learning which requires training in two stages. These contrastive approaches primarily focus on constructing unimodal invariance and equivariance or on designing pretext tasks, and the quality of these factors also influences the learned features. Diverging from these methods, we combine both by performing contrastive learning of cross-modal feature alignments and generating cross-modal mask patches as a means of self-supervised learning.

\subsection{{Cross-modal Methods}} 
Cross-modal methods involve pretraining models using data from different modalities, such as vision and language. 
By understanding the relationship among these modalities and facilitating the exchange of information between them, the model gains the ability to produce more robust representations, which are effective for various downstream tasks in the fields of vision and language \cite{radford2021learning, kim2021vilt, li2021align, lu2019vilbert, yao2021filip, yang2022vision}. 
The seminal work of CLIP \cite{radford2021learning} has demonstrated remarkable performance in predicting image-text correspondence, producing a task-agnostic model that can compete with supervised models trained for specific tasks. Building on this trend, recent works such as ALBEF \cite{li2021align} and TCL \cite{yang2022vision} focus on aligning features from different modalities before fusion, enhancing the capacity of cross-modal models to extract and utilize semantic information from heterogeneous sources. Additionally, several multi-modal methods emerged in the 3D field, either by mimicking the vision and language (VL) models or by using fixed image feature extractors to facilitate knowledge transfer.
CrossPoint \cite{DBLP:conf/cvpr/AfhamDDDTR22} learns point cloud representations using multi-modal contrastive loss. 
P2P \cite{wang2022p2p} performs cross-modal learning by converting 3D point clouds into 2D images using pretrained 2D models. 
CLIP2Point \cite{huang2022clip2point} and ACT \cite{dong2022autoencoders} leverage prior knowledge from pretrained VL models to benefit downstream tasks. 

Despite these efforts, the rigid nature of frozen VL models limits their flexibility. Effectively leveraging additional modal features to complement point cloud features and capitalizing on the strengths while addressing the limitations of self-supervised methods are crucial challenges for cross-modal self-supervised learning. 
\section{Method}
\label{sec:method}
Figure~\ref{fig:framework} presents the overall structure of Cross-BERT, a framework designed to learn point cloud representations from matched point-image pairs. Initially, we encode the augmented 3D point clouds and 2D images separately, generating two sequences of token embeddings which are then utilized in different ways. Firstly, we extract two fixed-size embeddings to calculate a fine-grained alignment score (Point-Image Alignment) between the input point cloud and image. Secondly, we input both sequences into the Interactive Cross-modal Encoder (ICE) module to obtain latent features through Masked Cross-modal Modeling (MCM). 

\subsection{Preprocessing of Point-image Pairs}

\textbf{Data Augmentation.} 
For an input point cloud, we apply various transformations, including scaling, rotation, and jittering, resulting in two augmented point clouds, denoted as $P$ and $P^{+}$.
To obtain the cross-modal input without using additional 2D data or specialized acquisition devices, we process the input object-level data and scene-level data in the following two ways, respectively. For object-level data, we generate two augmented images, denoted as $I$ and $I^{+}$, through two steps: (1) rendering the CAD model of point cloud $P$ from different angles to form pretraining image data, and (2) applying image augmentation techniques such as random cropping and color jittering. When dealing with scene-level data, we employ random cropping, resulting in two sub-scene images of the same size. We deliberately refrain from employing any supplementary image data augmentation techniques to maintain the intrinsic scene information.

Note that although both $P$ and $I$ undergo augmentation, we do not impose other symbols on them for convenience. In the subsequent pretraining process, we adopt a unified approach for both object-level and scene-level input pairs, eliminating the need to distinguish between the two.

\textbf{Patch Embeddings and Masking.} 
For the 3D modality, we use FPS (farthest point sampling) and KNN (k-nearest neighbor) algorithms to group points into patches. These patches are processed through mini-PointNet \cite{DBLP:conf/cvpr/QiSMG17} and linear projection to obtain embeddings, combined with positional embeddings. Similarly, we divide the 2D image into non-overlapping patches and generate image patch embeddings. Learnable class embeddings and positional embeddings are appended before sending the paired data to their respective encoders. 
It is worth noting that we do not intend to mask the images to maximize the supplementary information provided to the 3D model.

\subsection{Point Cloud and Image Encoders}
The deep representations extracted by transformer capture the interdependencies between the input patches and contain rich semantic information, making them suitable for refining supervision signals and various downstream tasks. 
Therefore, we utilize a 12-layer transformer \cite{vaswani2017attention} following the settings of Point-BERT \cite{DBLP:conf/cvpr/YuTR00L22} as the point cloud encoder and a 12-layer transformer \cite{dosovitskiy2020image} as the image encoder.
The point cloud representations generated by the point cloud encoder can be denoted as $\left\{z_{cls}^{P} \right\} \cup \left\{ z_{i}^{P} \right\}_{i=1}^{g}$. Herein $\left\{z_{cls}^{P} \right\}$ represents the learned point cloud class information, and for convenience, $\left\{ z_{i}^{P} \right\}_{i=1}^{g}$ denotes the representations for both masked and unmasked patches with the total number $g$. The masked patches can serve as the ground truth for computing the reconstruction loss during pretraining. Similarly, the image encoder produces the image representations $\left\{z_{cls}^{I} \right\} \cup \left\{ z_{i}^{I} \right\}_{i=1}^{v}$, which consist of the class information and the $v$ discrete encoded image features. 
In comparison to the analogous two-encoder method \cite{chen2023pimae}, our approach diverges in that it refrains from incorporating additional encoders to facilitate modality interactions. Instead, we endeavor to optimize cross-modal performance while concurrently mitigating computational overhead, drawing inspiration from the principles established in \cite{kim2021vilt}.

\subsection{Pretraining Tasks}
\subsubsection{Point-Image Alignment}
In the context of cross-modal learning, achieving sufficient cross-modal interactions plays a vital role in enhancing the representations of all modalities. 
However, the distinct embedding spaces of point cloud and image features pose a fundamental challenge in modeling their interactions, thus limiting the performance of these methods. 
To address it, we introduce an alignment task to align the two modalities into a shared embedding space to facilitate cross-modal interactions in the subsequent masked cross-modal modeling task.
The alignment task encompasses two contrastive objectives: Point-Image Contrastive (PIC) and Unimodal Contrastive (UMC), as shown in Figure~\ref{fig:framework}. The former facilitates the alignment of two distinct modalities into a shared embedding space, while the latter ensures the integrity of information within each modality.

\textbf{Point-Image Contrastive.}
\label{sec:pia}
PIC aims to minimize the distance between the embeddings of the matched point-image pair, which are sampled from the joint distribution while increasing the distance between the embeddings of the unmatched pairs, which are sampled from the product of marginal distributions. 
Inspired by MoCo \cite{KaimingHe2019MomentumCF} and SimCLR \cite{DBLP:conf/icml/ChenK0H20}, we maintain two queues to store the most recent $Q$ point-image representations from the momentum point cloud encoder and image encoder. 

For the pair $(P,I)$, the desired class embeddings of point cloud $P$ and image $I$ are represented as
$z^{P}_{cls}$ and $z^{I}_{cls}$, and their augmented editions $P^{+}$ and $I^{+}$ are denoted as $\bar{z}^{P}_{cls}$ and $\bar{z}^{I}_{cls}$, respectively. We define the similarity function $s\left ( P,I \right ) = f_{P} (z^{P}_{cls})^{T} f_{I} ({z}^{I}_{cls})$, where $f_{P}$ and $f_{I}$ are projection heads that map the representations to the same space. Then we define the softmax-normalized loss function as:
\begin{equation}
    L(X,Y) = \mathbb{E}_{p( X,Y )}  \frac{e^\frac{{s\left ( X,Y \right )}}{\tau }}{\sum_{q=1}^{Q} e^\frac{{s\left ( X,\widehat{Y_{q}} \right )}}{\tau }},
\end{equation}
where $\tau$ is the temperature parameter.
To maintain the negative samples $\widehat{Y}$, we use a large queue that keeps the most recent $Q$ projected representations.

Overall, the point-image contrastive loss is defined as:
\begin{equation}
    L_{pic}=\frac{1}{2}(L(P,I^{+}) + L(P^{+},I)).
\end{equation}
Intuitively, the minimization of $L_{pic}$ enables the alignment of 3D features and 2D features within the embedding space. This alignment, in turn, serves to facilitate the fusion of these features. To manage this process, we implement two queues of length $Q$ to store the projected representations of both the image and the point cloud.

\textbf{Unimodal Contrastive.}
Aligning the embeddings of the two modalities using the PIC loss alone may destroy the information contained in the original representation. To address this issue, we propose an Unimodal Contrastive (UMC) loss to encourage learning representations that maintain alignment between semantically-related positive pairs within each modality. For each point cloud and image, we calculate the point-to-point loss $L(P, P^{+})$ and image-to-image $L(I, I^{+})$ loss, respectively.
Subsequently, we minimize the unimodal contrastive loss to guarantee unimodal representation learning: 
\begin{equation}
L_{umc}=\frac{1}{2}(L(P,P^{+}) + L(I,I^{+})).
\end{equation}

PIC and UMC play complementary roles in representation learning: PIC brings together matched point-image pairs in the embedding space, while UMC maximizes agreement between differently augmented views of the same sample. By combining these two contrastive objectives, the quality of the learned representations is improved while facilitating joint cross-modal learning of the subsequent task. 

In contrast to CrossPoint \cite{DBLP:conf/cvpr/AfhamDDDTR22}, which similarly engages in contrastive learning between unimodal and cross-modal data, our Cross-BERT goes beyond merely minimizing the distance between global features. We extend contrastive learning to the class embeddings to facilitate unimodal and cross-modal interactions between all patches in the point clouds and images (since class embedding is related to all patch embeddings). Fundamentally, our approach focuses on alignment strategies that incorporate the Transformer architecture and is more suitable for dense prediction tasks than CrossPoint because it enhances the coherence and comprehensiveness of model understanding.

\subsubsection{Masked Cross-modal Modeling}
The traditional unimodal mask modeling approach treats mask prediction as a single-choice classification problem. It involves selecting the token ID with the highest probability from the probability distribution matrix generated by the tokenizer and assigning this token ID to the corresponding patch. However, when we employ a dVAE \cite{ramesh2021zero} as the tokenizer to generate the codebook, the tendency of dVAE to focus on fine-grained details can result in assigning the same token ID to patches with similar details but distinct semantics, as observed in \cite{peng2022beit}. This limits the fault tolerance rate of the single-choice approach. Furthermore, with the integration of cross-modal information, the task of predicting masked unimodal tokens for cross-modal representations becomes even more challenging. 

To address these issues, we introduce a Masked Cross-modal Modeling task based on  \cite{li2022mc,fu2023boosting}. This task involves (1) incorporating high-level semantic guidance into cross-modal mask prediction to alleviate the overemphasis on details by dVAE, and (2) converting single-choice predictions into multi-choice predictions to enhance the fault tolerance rate.

\textbf{Multi-choice.}
We employ a tokenizer with the same architecture as Point-BERT to encode sub-points of $P$
into discrete point tokens $\left\{ d_{i}\right\}_{i=1}^{g} \in V$, where $V$ represents the learned vocabulary codebook. Notably, unlike Point-BERT, which generates one-hot vectors during pre-training, we produce the probability distribution matrix without performing the ``argmax'' operation.

\textbf{Semantic Guidance.}
To enhance the tokenizer's comprehension of point cloud features to facilitate the masked modeling task, we undertake the following two endeavors: (1) conducting cross-modal interactions to generate semantically rich representations, and (2) utilizing the semantic information to guide the masked modeling task.

Firstly, we propose an interactive cross-modal encoder (ICE) consisting of self-attention, cross-attention, and feedforward modules. Specifically, given the intermediate point cloud and image representations $z_{i}^{P}$ and $z_{i}^{I}$ (without class embedding), we perform interactive cross-attention by sequentially employing the image representations and the point cloud representations as queries. The queries are updated taking into account the dependencies between the two modalities. This allows our model to effectively capture the intricate interaction patterns between the image and point cloud modalities, resulting in more comprehensive representations of the cross-modal data, denoted as $z_{i}^{PI}$. Note that our ICE is exclusively utilized during the training phase. 

To avoid excessive attention to fine-grained details of dVAE, we add additional semantic guidance to the masked modeling task using the representations produced by the ICE module. Specifically, we calculate the similarity between the representations of each patch and utilize this similarity to re-weight the probability distribution matrix $P(d)$. The similarity matrix $W\in R^{g\times g}$ is calculated as:
\begin{equation}
W_{i,k}=\frac{e^{\left(z^{PI}_{i}, z^{PI}_{k}\right)}}{\sum_{i=1}^{g}e^{\left<z^{PI}_{i}, z^{PI}_{k}\right>}},
\end{equation}
where $z^{PI}_{i}$ denotes the representations produced by the ICE module. Then, we obtain a refined probability distribution matrix $P'(d) = W \cdot P\left (  d\right )$. 

The final probability distribution $\hat{d}$ is a weighted sum of the original matrix $P\left ( d\right )$ and the refined matrix $P'(d)$:
\begin{equation}
\hat{d} = \alpha  P\left (  d\right ) + \left (  1-\alpha \right )P'\left (  d\right ),
\end{equation}
where $\alpha$ represents the weight introduced to balance between low-level semantics (directly predicted by the tokenizer) and high-level semantics.

\textbf{MCM Loss.}
Ultimately, we change the mask modeling objective from the original single-choice cross-entropy loss to a multi-choice cross-entropy loss formulated as:
\begin{equation}
    L_{mcm}=\mathbb{E}_{p(P^{msk},I)}\left [ -\sum_{k=1}^{\left| V\right|}\hat{d}_{i,k}\log h\left ( \hat{d}_{i,k}|z_{i}^{PI} \right ) \right ],
\end{equation}
where $\left| V\right|$ is the length of codebook vocabulary and $\sum_{k=1}^{\left| V\right|}\hat{d}_{i,k} = 1$. $h$ denotes an MLP head used for prediction.

The overall training objective of our model is given by:
\begin{equation}
     L= L_{mcm} + \beta L_{pic} + \zeta L_{umc},
\end{equation} 
where $\beta $ and $\zeta$ are weights balancing the different losses.
\section{Experiment}
\label{sec:experiment}

We proceed to assess the efficacy of the proposed Cross-BERT architecture across a diverse range of domains and tasks, encompassing both object- and scene-level scenarios.

\subsection{Experimental Settings}
\subsubsection{Discrete VAE details} 
Our dVAE comprises two main parts: a tokenizer and a decoder. The tokenizer is constructed using a 4-layer DGCNN (Dynamic Graph Convolutional Neural Network), while the decoder consists of a 4-layer DGCNN followed by FoldingNet \cite{DBLP:conf/cvpr/YangFST18}. FoldingNet generates the final output by combining 2D grids with input data.

To train the dVAE, we utilize the ShapeNet \cite{DBLP:journals/corr/ChangFGHHLSSSSX15} and SUN RGB-D \cite{song2015sun} datasets. The training specifics are outlined in Tables \ref{tab:hyper_params} and \ref{tab:hyper_params_sun}. Additionally, we establish a learnable vocabulary with a size of 8,192 tokens, and each token within this vocabulary is represented by a 256-dimensional vector.
\begin{table*}[h!]
\caption{Training recipes for dVAE, pretraining, and downstream finetuning for classification and segmentation.}
\label{tab:hyper_params}
\centering
\scalebox{1.0}{
\begin{tabular}{lccccc}
 & dVAE  & Pretraining & \multicolumn{2}{c}{Classification} & Segmentation\\
 \toprule[0.95pt]
 Config & ShapeNet & ShapeNet & ScanObjectNN & ModelNet & ShapeNetPart\\
 \midrule[0.6pt]
 optimizer & AdamW & AdamW & AdamW & AdamW & AdamW\\
 learning rate & 5e-4 & 5e-4 & 2e-5 & 1e-5 & 2e-4 \\
 weight decay & 5e-4 & 5e-2 & 5e-2 & 5e-2 & 5e-2 \\
 learning rate scheduler & cosine & cosine & cosine & cosine & cosine \\
 training epochs & 300 & 300 & 300 & 300 & 300\\
 warmup epochs & 10 & 10 & 10 & 10 & 10\\
 batch size & 64 & 128 & 32 & 32 & 16\\
 \midrule[0.6pt]
 number of points & 1024 & 1024 & 2048 & 1024/8192 & 2048 \\
 number of point patches & 64 & 64 & 128 & 64/512 & 128 \\
 point patch size & 32 & 32  & 32 & 32 & 32 \\
\bottomrule[0.95pt]
\end{tabular}
}
\end{table*}
\begin{table*}[h!]
\caption{Training recipes for dVAE, pretraining, and downstream finetuning for detection}.
\label{tab:hyper_params_sun}
\centering
\scalebox{1.0}{
\begin{tabular}{lccccc}
 & dVAE  & Pretraining & \multicolumn{3}{c}{Detection}\\
 \toprule[0.95pt]
 Config & SUN RGB-D & SUN RGB-D & SUN RGB-D & ScanNetV2 & KITTI\\
 \midrule[0.6pt]
 optimizer & AdamW & AdamW & AdamW & AdamW & AdamW\\
 learning rate & 5e-4 & 5e-4 & 1e-5 & 1e-5  & 2e-4 \\
 training epochs & 300 & 400 & 1080  & 1080  & 195\\
 warmup epochs & 10 & 10 & 9 & 9 & - \\
 batch size & 64 & 64 & 4 & 4 & 8\\
 \midrule[0.6pt]
 query & - & - & 128 & 256 & 50  \\
\bottomrule[0.95pt]
\end{tabular}
}
\end{table*}
\begin{table*}[t!]
\centering
\small
\footnotesize
\caption{
Classification results on ScanObjectNN and ModelNet40. Overall accuracy (\%) is reported. 
Cross-BERT is compared with the methods categorized as: $\circ$ methods using priori knowledge (\textit{e}.\textit{g}., pretrained image modal), $\bullet$ methods utilizing additional information (\textit{e}.\textit{g}., image), and $\diamond$ methods relying solely on geometric information.
}
\label{tab:scanobjectnn}
\resizebox{0.8\linewidth}{!}{
\begin{tabular}{cccccc}
\toprule[0.95pt]
\multirow{2}{*}[-0.5ex]{Method} & \multicolumn{3}{c}{ScanObjectNN} & \multicolumn{2}{c}{ModelNet40}\\
\cmidrule(lr){2-4}\cmidrule(lr){5-6} & OBJ\_BG & OBJ\_ONLY & PB\_T50\_RS& 1k P & 8k P\\
\midrule[0.6pt]
\multicolumn{6}{c}{Supervised Learning}\\
\midrule[0.6pt]
$\diamond$ PointNet~\cite{DBLP:conf/cvpr/QiSMG17}  & 73.3 & 79.2 & 68.0 & 89.2 & 90.8\\
$\diamond$ PointNet++~\cite{DBLP:conf/nips/QiYSG17}  & 82.3 & 84.3 & 77.9 & 90.7 & 91.9\\
$\diamond$ DGCNN~\cite{DBLP:journals/tog/WangSLSBS19}  & 82.8 & 86.2 & 78.1 & 92.9 & -\\
$\diamond$ PointCNN~\cite{DBLP:conf/nips/LiBSWDC18} & 86.1 & 85.5 & 78.5 & 92.2 & -\\
$\diamond$  PCT~\cite{guo2021pct} & - & - & - & 93.2 & -\\
$\diamond$ PointMLP~\cite{ma2022rethinking} & - & - & 85.4$\pm$0.3 & 94.5 & -\\
$\bullet$ 
P2P-RN101~\cite{wang2022p2p} & - & - & 87.4 & 93.1 & -\\
$\bullet$ 
P2P-HorNet~\cite{wang2022p2p} & - & - & 89.3 & 94.0 & -\\
\midrule[0.6pt]
\multicolumn{6}{c}{Self-Supervised Learning}\\
\midrule[0.6pt]

$\diamond$ Transformer + OcCo\cite{wang2021unsupervised}  & 84.85 & 85.54 & 78.7 & 92.1 & - \\ 
$\diamond$ 
Point-BERT~\cite{DBLP:conf/cvpr/YuTR00L22}  & 87.43 & 88.12 & 83.07 & 93.2 & 93.8\\
$\diamond$ MaskPoint~\cite{liu2022masked}  & 89.30 & 88.10 & 84.30 & 93.8 & -\\
$\diamond$ Point-MAE~\cite{pang2022masked}  & 90.02 & 88.29 & 85.18 & 93.8 & 94.0\\
$\diamond$ Point-M2AE~\cite{zhang2022point}  & 91.22 & 88.81 & 86.43 & 94.0 & -\\
$\bullet$ Joint-MAE~\cite{DBLP:journals/corr/abs-2302-14007}  & 90.94 & 88.86 & 86.0 & 94.0 & - \\
$\circ$ ACT~\cite{dong2022autoencoders} & 93.29 & 91.91 & 88.21 & 93.7 & 94.0\\
$\bullet$ Cross-BERT \textbf{w/o vot.} & \textbf{93.65} & \textbf{92.10} & \textbf{89.03} & \textbf{94.0} & \textbf{94.1}\\
$\bullet$ Cross-BERT~\textbf{w/ vot.} & \textbf{93.78} & \textbf{92.28} & \textbf{90.23} & \textbf{94.2} & \textbf{94.4}\\
\bottomrule[0.95pt]
\end{tabular}
}
\end{table*}

\begin{center}
\begin{table}[t!]
\vspace{-5pt}
\caption{Part segmentation on ShapeNetPart.  The mIoU over all classes and all instances are both reported. We use bold to indicate the highest accuracy and underline to indicate the next highest accuracy.
}
\label{tab:partseg}
\centering
\scalebox{1.0}{
\begin{tabular}{lcc}
\toprule[0.95pt]
Methods & mIoU (classes) (\%) & mIoU (instances) (\%)\\
\midrule[0.6pt]
$\diamond$ PointNet~\cite{DBLP:conf/cvpr/QiSMG17}  & 80.4 & 83.7\\
$\diamond$ PointNet++~\cite{DBLP:conf/nips/QiYSG17} & 81.9 & 85.1\\
$\diamond$ DGCNN~\cite{DBLP:journals/tog/WangSLSBS19} & 82.3 & 85.2\\
\midrule[0.6pt]
$\diamond$ PointMLP~\cite{ma2022rethinking} & 84.6 & 86.1\\
$\diamond$ Transformer~\cite{DBLP:conf/iccv/ZhaoJJTK21} & 83.4 & 84.7\\
$\diamond$ PointContrast~\cite{xie2020pointcontrast} & - & 85.1\\
$\bullet$  CrossPoint~\cite{DBLP:conf/cvpr/AfhamDDDTR22} & - & 85.5\\
$\diamond$ 
Point-BERT~
\cite{DBLP:conf/cvpr/YuTR00L22} & 84.1 & 85.6\\
$\diamond$ 
Point-MAE~\cite{pang2022masked} & - & 86.1\\
$\circ$ 
ACT~\cite{dong2022autoencoders} & 84.7 & 86.1\\
$\bullet$ 
Joint-MAE~\cite{DBLP:journals/corr/abs-2302-14007} & \textbf{85.4} & 86.3 \\
$\bullet$ Cross-BERT & \underline{84.8} & \textbf{86.4} \\
\bottomrule[0.95pt]
\end{tabular}
}
\end{table}
\end{center}
\subsubsection{Encoder Details}
The point cloud encoder in our approach follows the well-established Transformer architecture \cite{vaswani2017attention}. It consists of stacked Transformer blocks, where each block is composed of a multi-head self-attention layer and a FeedForward Network (FFN). LayerNorm (LN) is applied to both layers to normalize the outputs. Similarly, we utilize the Vision Transformer (ViT) \cite{dosovitskiy2020image} as the image encoder.

To enable Interactive Multimodal Encoder (IME), we utilize two transformer blocks. Each block includes a self-attention layer, a cross-attention layer, and an FFN. The cross-attention layer in the first block takes the image tokens as Q (query) and the point cloud tokens as KV (key and value) to enhance the image tokens. Conversely, the cross-attention layer in the second block employs the point cloud tokens as Q and the tokens of the image modality as KV. 

For the ShapeNet dataset, the hidden dimension of each Transformer block is set to 384, and the number of heads in each self-attention layer and cross-attention layer is set to 6 and 12, respectively. For the SUN RGB-D dataset, the hidden dimension of each Transformer block is set to 256, and the number of heads of the self-attention layer is adjusted to 4. 

For further details on the architectures and training configurations, please refer to Tables \ref{tab:hyper_params} and \ref{tab:hyper_params_sun}.

\subsubsection{Pretraining setting}
To accomplish the evaluation, we undertake a pretraining phase utilizing two distinct datasets.

For object-level experiments, we follow the previous works~\cite{pang2022masked,zhang2022point}, to pretrain our model on ShapeNet \cite{DBLP:journals/corr/ChangFGHHLSSSSX15}.ShapeNet covers over 51,000 unique 3D models from 55 common object categories. We sample surface points from the 3D object models to create 3D point clouds. 1,024 points are uniformly sampled from each 3D model.
Additionally, we employ texture lighting to render high-quality 2D RGB images. We use the rendered images from~\cite{huang2022clip2point}. These images consist of 10 diverse views for each point cloud, and from this set, we randomly select two images. During pretraining, we conduct 300 epochs using a cosine learning rate schedule with a learning rate of 5e-4. The learning rate is warmed up for 10 epochs. We employ the AdamW optimizer \cite{DiederikPKingma2014AdamAM} and use a batch size of 128.

For scene-level ones, we adopt the setting outlined in \cite{chen2023pimae}, leveraging both the supplied image data and the point cloud data generated from the SUN RGB-D dataset \cite{song2015sun}. SUN RGB-D is a challenging large-scale 3D indoor dataset, comprising 10,335 RGB-D images. This dataset includes meticulously labeled 3D bounding boxes spanning 37 distinct categories. The conversion of depth images to point clouds is accomplished through the utilization of provided camera poses. Our work aligns with the established convention, employing the standard 5,285-image training split and a separate 5,050-image split for the testing phase.
After pretraining, we append different heads onto the point cloud encoder for different downstream tasks. For each 3D scene in SUN RGB-D, we use a single set aggregation operation to subsample \cite{DBLP:conf/cvpr/QiSMG17} 2048 points and obtain 256-dimensional point features. During pretraining, we conduct 400 epochs using a cosine learning rate schedule with a learning rate of 1e-3. The learning rate is warmed up for 15 epochs. We employ the AdamW optimizer \cite{DiederikPKingma2014AdamAM} and use a batch size of 64.

Further details can be found in Table \ref{tab:hyper_params}. The hyperparameters $\alpha$, $\beta$, and $\zeta$ in MCM Loss are set to 0.8, 1.0, and 1.0, respectively.

\subsubsection{Downstream Task Details}
During the transfer to downstream tasks, only the point cloud encoder from the Cross-BERT model is utilized for classification, segmentation, and detection tasks.

\textbf{Classification task} We employ the following approach: the feature of the class token is extracted, and max pooling is applied to the remaining token features. These pooled features are then combined and fed into a two-layer MLP with dropout for classification. To evaluate the performance of our approach, we conducted experiments on two datasets: ModelNet40 \cite{DBLP:conf/cvpr/WuSKYZTX15} and ScanObjectNN \cite{DBLP:conf/iccv/UyPHNY19}.
ModelNet40 is a widely used dataset for object classification, consisting of 12,331 meshed 3D CAD models covering 40 different categories. It provides a benchmark for evaluating classification performance on clean and well-structured 3D CAD models. ScanObjectNN is a dataset that presents a more challenging scenario for object classification. It contains 2,880 3D object point clouds collected from real-world indoor scenes, spanning 15 categories. This dataset is designed to assess the performance of classification algorithms on point clouds obtained from complex and cluttered environments.

\textbf{Segmentation task} We use the ShapeNetPart dataset~\cite{DBLP:journals/corr/ChangFGHHLSSSSX15}, which contains 16,881 3D objects from 16 categories, annotated with 50 parts in total. Our method aligns with Point-BERT, where we leverage the learned features from specific layers of the Transformer block. More specifically, we utilize the features from the 4th, 8th, and 12th layers. These features are concatenated, and both average pooling and max pooling are applied to obtain two global features. Additionally, the concatenated features represent 128 center points, and they are up-sampled to 2,048 points to obtain per-point features. These per-point features are then concatenated with the two global features, and an MLP is applied to predict the label for each point.

\textbf{Detection task} 
We evaluate our method on two datasets: SUN RGB-D and ScanNetV2 \cite{dai2017scannet}. ScanNetV2 is a comprehensive 3D interior scene dataset featuring 1,513 indoor scenes and covering 18 object classes. The dataset comes with rich annotations, including semantic labels, per-point instances, and 2D as well as 3D bounding boxes. For assessing 3D object detection, we follow standard evaluation metrics \cite{demisra2021end}, specifically measuring mean Average Precision (mAP) at IoU thresholds of 0.25 and 0.5. In our evaluation, we employ 3DETR as the baseline and subsequently substitute 3DETR's encoder with the pretrained point cloud encoder from Cross-BERT. The fine-tuning process is executed while adhering rigorously to the same settings as the original \cite{demisra2021end}.

More details on the architecture of the classification and segmentation heads are provided in Table \ref{tab:hyper_params} and Table \ref{tab:hyper_params_sun}.

\subsection{Downstream Tasks}
\subsubsection{Object Classification}

For the synthetic object classification experiments, we utilize the ModelNet40 \cite{DBLP:conf/cvpr/WuSKYZTX15} dataset. We compare our method with several representative supervised methods, as well as state-of-the-art self-supervised uni- and cross-modal pretraining methods. 
We adopt classification accuracy as the evaluation metric, and the results are reported in Table \ref{tab:scanobjectnn}. As observed, our method obtains 94.4\% accuracy on ModelNet40, outperforming its competitors and achieving new state-of-the-art performance.

For the real-world object classification experiments, we use the challenging ScanObjectNN \cite{dai2017scannet} dataset, which is made up of occluded objects captured from real-world indoor scenes. We use the same settings as \cite{DBLP:conf/cvpr/QiSMG17, DBLP:journals/tog/WangSLSBS19} for fine-tuning. We conduct experiments on three main variants: OBJ-BG, OBJ-ONLY, and PB-T50-RS. The experimental results are reported in Table \ref{tab:scanobjectnn}, where Cross-BERT outperforms the state-of-the-art self-supervised methods, proving the generality of our proposed method for out-of-distribution data. Compared to the previous SOTA ACT \cite{dong2022autoencoders} and Joint-MAE \cite{DBLP:journals/corr/abs-2302-14007}, our performance, while only slightly improved, spans the original effectiveness gap between BERT-style methods and MAE-based methods. 

\begin{center}
\begin{table}[t!]
    \caption{Few-shot classification results on ModelNet40. The overall accuracy (\%) under different setting without voting are reported.
}
    \label{tab:few-shot}
    \centering
\resizebox{0.7\linewidth}{!}{
    \begin{tabular}{lcccc}
    \toprule[0.95pt]
    \multirow{2}{*}[-0.5ex]{Method}& \multicolumn{2}{c}{5-way} & \multicolumn{2}{c}{10-way}\\
    \cmidrule(lr){2-3}\cmidrule(lr){4-5} & 10-shot & 20-shot & 10-shot & 20-shot\\
    \midrule[0.6pt]
    $\diamond$ 
    DGCNN~\cite{DBLP:journals/tog/WangSLSBS19} &31.6 $\pm$ 2.8 &  40.8 $\pm$ 4.6&  19.9 $\pm$  2.1& 16.9 $\pm$ 1.5\\
    $\diamond$
    Transformer~\cite{DBLP:conf/iccv/ZhaoJJTK21} & 87.8 $\pm$ 5.2& 93.3 $\pm$ 4.3 & 84.6 $\pm$ 5.5 & 89.4 $\pm$ 6.3\\
    $\diamond$ OcCo~\cite{wang2021unsupervised} & 94.0 $\pm$ 3.6& 95.9 $\pm$ 2.3 & 89.4 $\pm$ 5.1 & 92.4 $\pm$ 4.6\\
    $\diamond$ Point-BERT~\cite{DBLP:conf/cvpr/YuTR00L22} & 94.6 $\pm$ 3.1 & 96.3 $\pm$ 2.7 &  91.0 $\pm$ 5.4 & 92.7 $\pm$ 5.1\\
    $\diamond$ MaskPoint~\cite{liu2022masked} & 95.0 $\pm$ 3.7 & 97.2 $\pm$ 1.7 & 91.4 $\pm$ 4.0 & 93.4 $\pm$ 3.5\\
    $\diamond$ Point-MAE~\cite{pang2022masked} & 96.3 $\pm$ 2.5&97.8 $\pm$ 1.8 & 92.6 $\pm$ 4.1 & 95.0 $\pm$ 3.0\\
    $\diamond$ Point-M2AE~\cite{zhang2022point} & 96.8 $\pm$ 1.8&  \textbf{98.3 $\pm$ 1.4} & 92.3 $\pm$ 4.5 & 95.0 $\pm$ 3.0\\
    $\circ$ ACT~\cite{dong2022autoencoders} & 96.8 $\pm$ 2.3 & 98.0 $\pm$ 1.4 & \textbf{93.3 $\pm$ 4.0} & 95.6 $\pm$ 2.8\\
    $\bullet$ Joint-MAE~\cite{DBLP:journals/corr/abs-2302-14007} & 96.7 $\pm$ 2.2 & 97.9 $\pm$ 1.8 & 92.6 $\pm$ 3.7 & 95.1 $\pm$ 2.6 \\
    $\bullet$ Cross-BERT & \textbf{97.0 $\pm$ 2.1} & \underline{98.2 $\pm$ 1.3} & \underline{93.0 $\pm$ 3.4} & \textbf{95.6 $\pm$ 3.0} \\
    
    \bottomrule[0.95pt]
    
    \end{tabular}
}
\end{table}
\end{center}
\begin{table}[h]
\centering
\caption{3D object detection results on ScanNetV2 \cite{dai2017scannet} and SUN RGB-D~\cite{song2015sun}.We adopt the average precision with 3D IoU thresholds of 0.25 (AP25) and 0.5 (AP50) for the evaluation metrics.}
\label{tab:detection}

\resizebox{0.6\linewidth}{!}{
\begin{tabular}{l|ccccc}

\toprule
& & \multicolumn{2}{c}{SUN RGB-D} & \multicolumn{2}{c}{ScanNetV2} \cr
Methods & Pretrained & $AP_{25}$ & $AP_{50}$ &  $AP_{25}$ & $AP_{50}$\\
\midrule
PointFusion \cite{xu2018pointfusion} & \textit{None} & 45.4 & - & - & - \\
3D-SIS \cite{hou20193d} & \textit{None} & - & - & 40.2 & 22.5 \\

VoteNet \cite{qi2019deep} & \textit{None} & 57.7 & 32.9 & 58.6 & 33.5 \\
\midrule
3DETR \cite{demisra2021end} & \textit{None} & 58.0 & 30.3 & 62.1 & 37.9 \\
+PiMAE \cite{chen2023pimae} & SUN RGB-D & \textbf{59.4} & 33.2 & 62.6 & \textbf{39.4} \\
+Cross-BERT & SUN RGB-D & \underline{58.9} & \textbf{33.7} & \textbf{62.7} & \textbf{39.4} \\
\bottomrule

\end{tabular}
}
\end{table}

\subsubsection{Object Part Segmentation}
For 3D object part segmentation, we choose ShapeNetPart \cite{DBLP:journals/tog/YiKCSYSLHSG16}. We follow the setting in PointNet \cite{DBLP:conf/cvpr/QiSMG17} and randomly sample 2,048 points from each model. We compare our method with the widely used supervised methods 
and the recently published self-supervised pretraining methods. 
The mIoU results are reported in Table~\ref{tab:partseg}, which clearly shows that our Cross-BERT outperforms supervised methods~\cite{DBLP:conf/cvpr/QiSMG17,DBLP:conf/nips/QiYSG17,DBLP:journals/tog/WangSLSBS19}. Specifically, our method achieves the best performance in mIoU (instances), outperforming the most advanced self-supervised techniques.

\subsubsection{Object Few-shot Learning}
We follow \cite{sharma2020self} to evaluate our model's performance in few-shot learning. We use the ``K-way N-shot" setting and randomly select $K$ classes and sample $(N+20)$ objects for each class. The model is trained on the support set, which consists of $K \times N$ samples, and evaluated on the remaining 20K samples in the query set. We conduct experiments under different settings including ``5way 10shot'', ``5way 20shot'', ``10way 10shot'', and ``10way 20shot''. 
We then report the average performance and standard deviation over 10 runs in Table \ref{tab:few-shot}. The results show that Cross-BERT produces higher accuracy than SOTA methods in almost all few-shot settings. These results demonstrate the strong capacity of our model with limited data.

\subsubsection{Indoor 3D Object Detection}
We employ our 3D feature extractor to enhance the performance of 3D detectors, achieved by either replacing or inserting the encoder into different backbone architectures. 
We utilize 3DETR \cite{demisra2021end} as the baseline, and the results are presented in Table \ref{tab:detection}, demonstrating substantial enhancements brought about by our model. For evaluation purposes, we adopt the metrics presented in \cite{chen2023pimae}, measuring the Average Precision (AP) of 3D bounding boxes. AP25 and AP50 use 0.25 and 0.5 as thresholds for IoU, respectively. Our approach yields superior or sub-superior results in all metrics. Meanwhile, our method works better compared to the original 3DERT, showing the effectiveness of the pretraining process.

\subsubsection{Outdoor 3D Object Detection}

Our evaluation extends to challenging outdoor scene datasets, specifically the KITTI dataset \cite{geiger2012we}, characterized by significantly distinct data distribution compared to indoor data. The KITTI dataset is a widely recognized benchmark for outdoor 3D object detection, comprising 7,481 training images and 7,518 test images. We follow the metrics presented in \cite{chen2023pimae}, measuring the AP of 3D bounding boxes across three difficulty levels: easy, moderate, and hard. As presented in Table \ref{tab:mono3d}, our method exhibits competence over easy and moderate levels and improvements over hard levels in MonoDETR \cite{zhang2022monodetr}. This outcome serves as empirical evidence affirming our model's adeptness in generalizing pretrained representations across diverse datasets. Moreover, in comparison to PiMA~\cite{chen2023pimae}, which shares the same two-encoder structure, Cross-BERT achieves similar or even higher accuracy, thus validating the efficacy of our approach.

\subsubsection{Few-shot Classification of Images}
\begin{table}[t]
\caption{Few-shot image classification results on CIFAR-FS. We show top-1 classification accuracy under 5-way 1-shot and 5-way 5-shot settings.} 
\label{tab:imagefs}
\centering
\resizebox{0.5\linewidth}{!}{
\begin{tabular}{lccc}

\toprule[0.95pt]
Method & backbone & 5-way 1-shot & 5-way 5-shot\\
\midrule[0.6pt]
CrossPoint~\cite{DBLP:conf/cvpr/AfhamDDDTR22} & ResNet50 &  64.5 & 80.1\\
PiMAE~\cite{chen2023pimae}& ViT& 66.9 & \textbf{80.7} \\
Cross-BERT (ours)  & ViT & \textbf{67.9} & 80.5 \\
\bottomrule[0.95pt]

\end{tabular} }
\end{table}
In addition to the primary point cloud processing tasks, we also conducted a simple image classification experiment to explore the capability of our model in image understanding. Following CrossPoint \cite{DBLP:conf/cvpr/AfhamDDDTR22}, we utilized the CIFAR-FS dataset \cite{tian2020rethinking}, which is a commonly used dataset for few-shot image classification. CIFAR-FS consists of 100 categories with 64, 16, and 20 samples for the train, validation, and test sets, respectively. We use no extra design for the classifier, only adding a linear layer to the feature encoder, predicting the category based on class embedding as input. The results are presented in Table \ref{tab:imagefs}. While our model did not achieve exceptionally high accuracy in this classification task, the findings provide evidence that the image encoder within our framework is capable of learning certain features from real images through cross-domain pretraining.

\begin{table} [t]
\makeatletter\def\@captype{table}
\caption{Monocular \cite{zhang2022monodetr} 3D object detection results of car category on KITTI\cite{geiger2012we} set. We adjust Monocular encoder to use PiMAE's depth encoder, loading the pretrained weights of PiMAE \cite{chen2023pimae} and Cross-BERT into the encoder.}
\centering
\scalebox{1.0}{

\begin{tabular}{lccc} 
    \toprule
     Methods & Easy & Mod. & Hard \\
    \midrule
    MonoDETR + PiMAE \cite{chen2023pimae} & \textbf{26.6} & \textbf{18.8} & 15.5 \\ 
    MonoDETR + Cross-BERT & 26.5 & 18.5 & \textbf{15.7} \\ 
    \bottomrule
  \end{tabular}
  }

\label{tab:mono3d}
\end{table}

\begin{table}[t!]
\caption{Ablation study on pretraining targets. Overall accuracy (\%) is linear SVM is reported.} \label{tab:target}
\centering
\resizebox{0.5\linewidth}{!}{

\begin{tabular}{ccccc}
\toprule[0.95pt]
\multicolumn{2}{c}{Mask modeling} & \multicolumn{2}{c}{Alignment} & \multirow{2}{*}[-0.5ex]{ModelNet40} \\
\cmidrule(lr){1-2} \cmidrule(lr){3-4} Multi-choice & ICE  &  PIC & UMC \\
\midrule[0.6pt] 
$\times$ & $\times$ & $\times$ & $\checkmark$ &  88.9\\
$\times$  & $\times$ & $\checkmark$ & $\times$ & 89.6 \\
$\times$ & $\times$ & $\checkmark$ & $\checkmark$ & 90.0 \\
\midrule[0.6pt] 
$\checkmark$ & $\times$ & $\checkmark$ & $\checkmark$ & 90.9 \\
$\times$ & $\checkmark$ &$\checkmark$ & $\checkmark$ & 90.4 \\
$\checkmark$ & $\checkmark$ & $\checkmark$ & $\checkmark$ &  91.5 \\
\bottomrule[0.95pt]
\end{tabular}
}
\end{table}

\subsection{Ablation Study}

\begin{figure}[t]
    \begin{center}

    \includegraphics[width=0.9\linewidth]{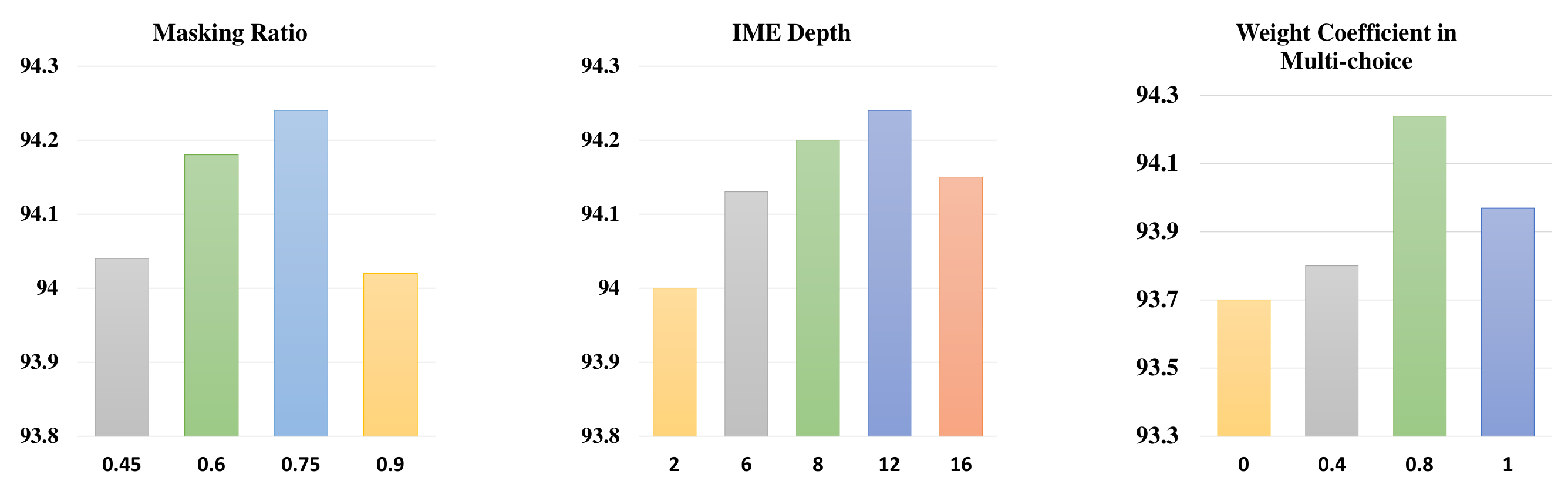}
    
    \caption{Ablation study of masking ratio, IME depth, and the weight coefficient in multi-choice used during Cross-BERT pretraining. The results is setting without voting.}.
    
    \label{fig:abhy}
    \end{center}
\end{figure}

\begin{figure}[!t]
    \begin{center}
    \includegraphics[width=0.9\linewidth]{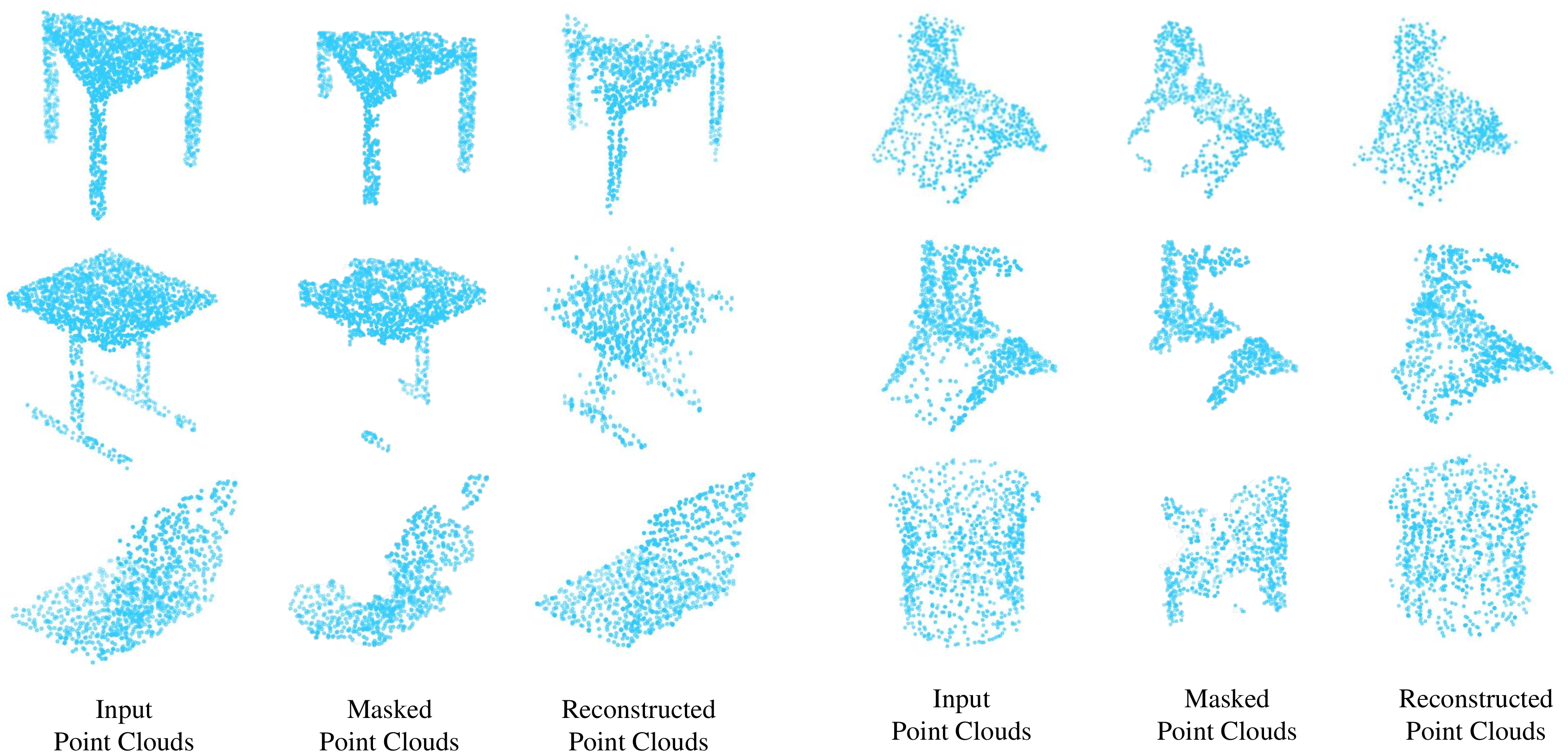}
    \caption{Reconstruction results on ShapeNet. From left to right: input point clouds (\textbf{i}.\textbf{e}., ground truths), masked point clouds, and reconstructed point clouds. The masking ratio was set to 45\%. }
    \label{fig:res}
    \end{center}
\end{figure}

\begin{figure}[!t]
    \begin{center}
    \includegraphics[width=0.9\linewidth]{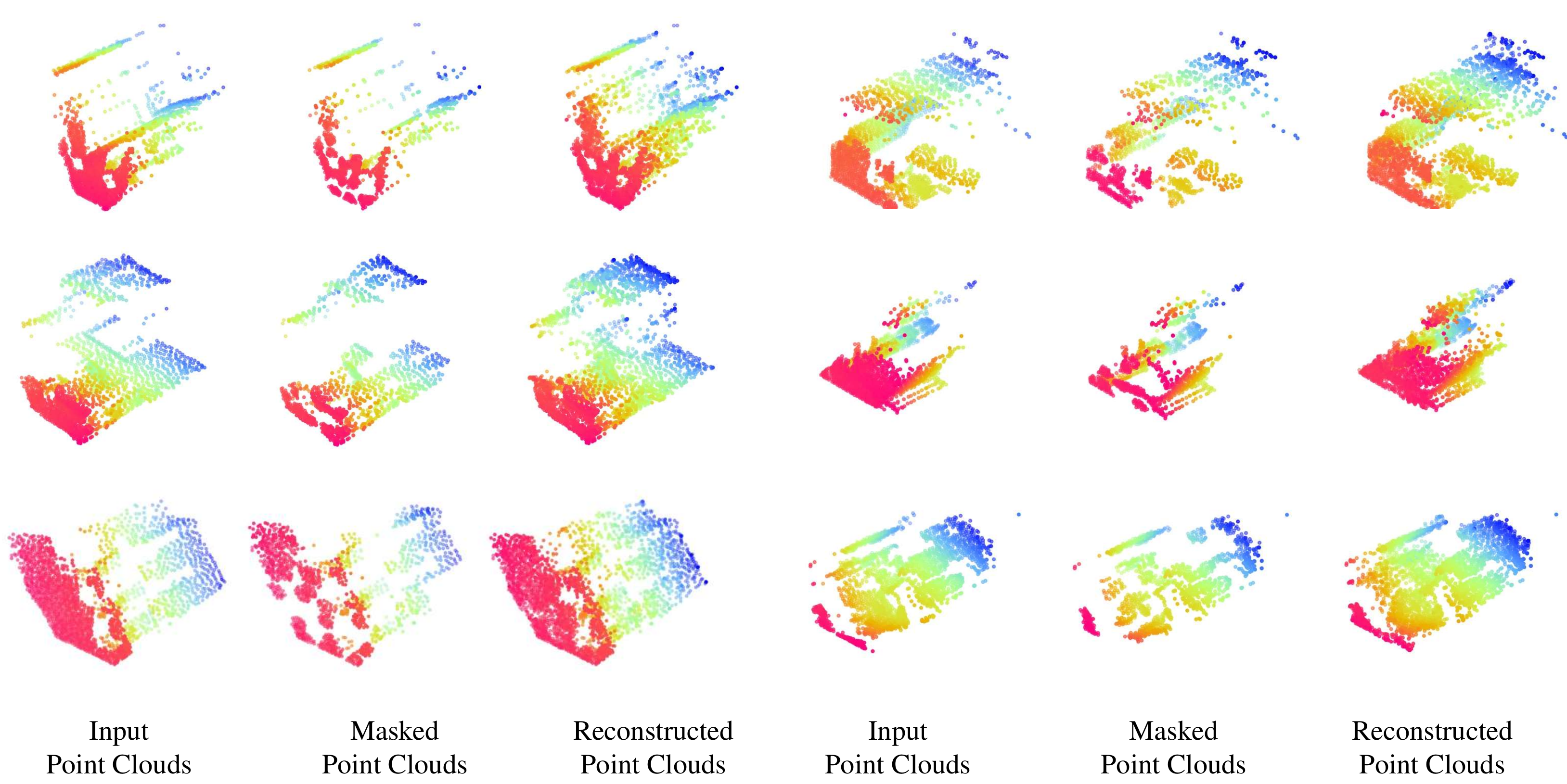}
    \caption{Reconstruction results on Sun RGB-D. From left to right: input point clouds (\textbf{i}.\textbf{e}., ground truths), masked point clouds, and reconstructed point clouds.}
    \label{fig:res_scene}
    \end{center}
\end{figure}

\begin{figure*}[h!]
    \begin{center}
    \includegraphics[width=1\linewidth]{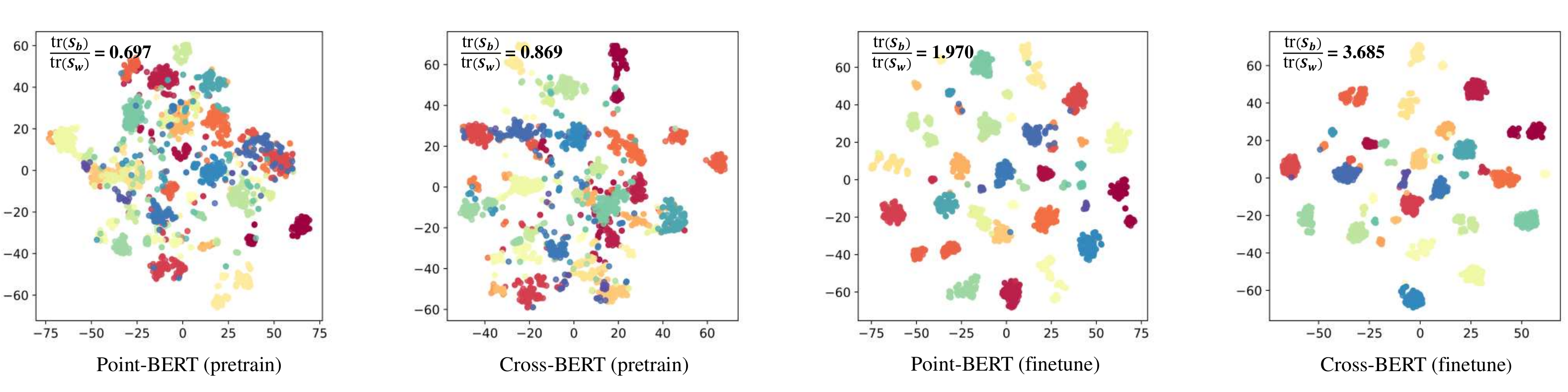}
     \caption{Visual comparison between t-SNE feature embeddings of Point-BERT and Cross-BERT. $\frac{tr(S_{b})}{tr(S_{w})}$ indicates the quality of the learned features (the higher the better).
    }
    \label{fig:tsne}
    \end{center}
\end{figure*}

\subsubsection{Ablation on Components}
To fully dissect the contributions of key components in Cross-BERT, we conduct thorough ablation experiments on ModelNet40, and for convenience, we evaluate the resulting performance based on the linear SVM classification accuracy. Specifically, we freeze the pre-trained point cloud feature extractor and use a simple linear SVM classifier on the training subset of the classification dataset.

We explore the effects of mask modeling strategies and alignment strategies. It is important to note that when solely utilizing the PIC strategy and not the ICE strategy, it signifies that we are exclusively employing image information to align cross-modal global tokens, without utilizing this information to predict the masked token ID. On the other hand, if only the UMC strategy is utilized, it indicates that we do not employ RGB images for the pretraining process, essentially representing the original Point-BERT configuration.

The results are presented in Table \ref{tab:target}, from which we can conclude that: (1) When mask modeling is not utilized, the model's performance is severely hindered by the overfitting of limited 3D data and solely relying on contrastive learning; (2) The results are even less effective when only PIC is used than when no image modality is introduced, but the combination of the two leads to better learning capabilities; (3) The alignment task plays a more important role than the mask modeling task. This is because the aligned cross-modal features simultaneously facilitate the mask modeling process, and the combination of the two tasks further enhances the ability to learn discriminative features.  

\subsubsection{Ablation on Hyperparameters}

We also demonstrate ablation studies of several significant hyperparameters.
Figure \ref{fig:abhy} presents an ablation study on \textit{masking ratio}, \textit{decoder depth}, and the \textit{weight coefficient} in multi-choice during the pretraining phase of Cross-BERT. The results indicate that an appropriate masking ratio and deeper IME depth contribute to improved representation learning capabilities. Regarding the weight coefficient $\alpha$, larger values tend to yield better pretraining performance. However, when $\alpha = 1.0$, the supervision signals do not incorporate high-level semantics, which is similar to Point-BERT. In contrast, the option of \textit{weight coefficient} plays a greater role. Both the choice of hyperparameters and the generation of high-level semantics by the transformer influence the pretraining.

\subsection{Visualization}
\subsubsection{Reconstruction Result}
In Figure~\ref{fig:res} and Figure~\ref{fig:res_scene}, we demonstrate some reconstruction results of our proposed method. We utilize the pretrained Cross-BERT model to generate a sequence of token IDs for the masked point cloud, which is then used for point cloud reconstruction. The results demonstrate the effectiveness of our model in reconstructing point cloud shapes and scenes, even when the complete shape of the object is not visible. Notably, the reconstructed point cloud successfully restores the covered corner portion of the laptop point cloud shown in the second row (left), despite the border being obstructed. 

\subsubsection{T-SNE Result}
Figure \ref{fig:tsne} presents a visual comparison between the t-SNE embeddings of Cross-BERT and Point-BERT, both pretrained on ShapeNet and finetuned on ModelNet40. Several observations can be made from this comparison: (1) Even when pretrained only on ShapeNet, our model produces discriminative features. (2) Finetuning further enhances the features. (3) Compared to Point-BERT, our features exhibit higher within-class similarity (indicated by a higher aggregation of the points in the same color) and greater between-class dissimilarity (as different sets of colored points are farther apart). We also employed the $\frac{tr(S_{b})}{tr(S_{w})}$ measure to quantitatively assess the quality of the learned features, where $S_b$ denotes between-class scatter matrix, $S_w$ denotes the within-class scatter matrix, and $tr(\cdot)$ computes the trace of a matrix. This measure supports the superiority of our representations, which aligns with the visual comparison.
\section{Conclusion}
We propose Cross-BERT, a cross-modal framework for 3D point cloud object and scene pretraining, which uses two specially tailored multimodal tasks to facilitate 3D representation learning from different perspectives. Our results not only demonstrate high-capacity data efficiency and generalization capabilities in both pretraining and downstream representation transferring tasks but also cross the long-standing effectiveness gap between BERT-based and MAE-based approaches in point cloud pretraining.
This self-supervised approach offers great potential for transferring cross-modal knowledge, which opens new avenues for advancing foundational modeling in the field of data-driven deep learning.
While Cross-BERT shows promising results, there are still areas that require further exploration. For example, additional experiments can be conducted to investigate the impact of different image rendering angles on feature learning. Furthermore, optimizing the performance by studying the selection of the number of images employed in pretraining is an avenue worth pursuing. Simultaneously, with the popularity of large-scale models, the question of how to conveniently and effectively leverage the capabilities of visual large models, language large models, or even large-scale models of other modalities to aid 3D representation learning is also an urgent one.

\bibliographystyle{ACM-Reference-Format}
\bibliography{sample}


\begin{thebibliography}{60}


\ifx \showCODEN    \undefined \def \showCODEN     #1{\unskip}     \fi
\ifx \showDOI      \undefined \def \showDOI       #1{#1}\fi
\ifx \showISBNx    \undefined \def \showISBNx     #1{\unskip}     \fi
\ifx \showISBNxiii \undefined \def \showISBNxiii  #1{\unskip}     \fi
\ifx \showISSN     \undefined \def \showISSN      #1{\unskip}     \fi
\ifx \showLCCN     \undefined \def \showLCCN      #1{\unskip}     \fi
\ifx \shownote     \undefined \def \shownote      #1{#1}          \fi
\ifx \showarticletitle \undefined \def \showarticletitle #1{#1}   \fi
\ifx \showURL      \undefined \def \showURL       {\relax}        \fi
\providecommand\bibfield[2]{#2}
\providecommand\bibinfo[2]{#2}
\providecommand\natexlab[1]{#1}
\providecommand\showeprint[2][]{arXiv:#2}

\bibitem[Achlioptas et~al\mbox{.}(2018)]%
        {achlioptas2018learning}
\bibfield{author}{\bibinfo{person}{Panos Achlioptas}, \bibinfo{person}{Olga Diamanti}, \bibinfo{person}{Ioannis Mitliagkas}, {and} \bibinfo{person}{Leonidas Guibas}.} \bibinfo{year}{2018}\natexlab{}.
\newblock \showarticletitle{Learning representations and generative models for {3D} point clouds}. In \bibinfo{booktitle}{\emph{ICML}}. \bibinfo{pages}{40--49}.
\newblock


\bibitem[Afham et~al\mbox{.}(2022)]%
        {DBLP:conf/cvpr/AfhamDDDTR22}
\bibfield{author}{\bibinfo{person}{Mohamed Afham}, \bibinfo{person}{Isuru Dissanayake}, \bibinfo{person}{Dinithi Dissanayake}, \bibinfo{person}{Amaya Dharmasiri}, \bibinfo{person}{Kanchana Thilakarathna}, {and} \bibinfo{person}{Ranga Rodrigo}.} \bibinfo{year}{2022}\natexlab{}.
\newblock \showarticletitle{{CrossPoint}: Self-Supervised Cross-Modal Contrastive Learning for {3D} Point Cloud Understanding}. In \bibinfo{booktitle}{\emph{CVPR}}. \bibinfo{pages}{9892--9902}.
\newblock


\bibitem[Bao et~al\mbox{.}(2022)]%
        {bao2021beit}
\bibfield{author}{\bibinfo{person}{Hangbo Bao}, \bibinfo{person}{Li Dong}, \bibinfo{person}{Songhao Piao}, {and} \bibinfo{person}{Furu Wei}.} \bibinfo{year}{2022}\natexlab{}.
\newblock \showarticletitle{{BEiT}: {BERT} Pre-Training of Image Transformers}. In \bibinfo{booktitle}{\emph{ICLR}}.
\newblock


\bibitem[Chang et~al\mbox{.}(2015)]%
        {DBLP:journals/corr/ChangFGHHLSSSSX15}
\bibfield{author}{\bibinfo{person}{Angel~X. Chang}, \bibinfo{person}{Thomas~A. Funkhouser}, \bibinfo{person}{Leonidas~J. Guibas}, \bibinfo{person}{Pat Hanrahan}, \bibinfo{person}{Qi{-}Xing Huang}, \bibinfo{person}{Zimo Li}, \bibinfo{person}{Silvio Savarese}, \bibinfo{person}{Manolis Savva}, \bibinfo{person}{Shuran Song}, \bibinfo{person}{Hao Su}, \bibinfo{person}{Jianxiong Xiao}, \bibinfo{person}{Li Yi}, {and} \bibinfo{person}{Fisher Yu}.} \bibinfo{year}{2015}\natexlab{}.
\newblock \showarticletitle{{ShapeNet}: An Information-Rich {3D} Model Repository}.
\newblock \bibinfo{journal}{\emph{CoRR}}  \bibinfo{volume}{abs/1512.03012} (\bibinfo{year}{2015}).
\newblock


\bibitem[Chen et~al\mbox{.}(2023)]%
        {chen2023pimae}
\bibfield{author}{\bibinfo{person}{Anthony Chen}, \bibinfo{person}{Kevin Zhang}, \bibinfo{person}{Renrui Zhang}, \bibinfo{person}{Zihan Wang}, \bibinfo{person}{Yuheng Lu}, \bibinfo{person}{Yandong Guo}, {and} \bibinfo{person}{Shanghang Zhang}.} \bibinfo{year}{2023}\natexlab{}.
\newblock \showarticletitle{Pimae: Point cloud and image interactive masked autoencoders for 3d object detection}. In \bibinfo{booktitle}{\emph{Proceedings of the IEEE/CVF Conference on Computer Vision and Pattern Recognition}}. \bibinfo{pages}{5291--5301}.
\newblock


\bibitem[Chen et~al\mbox{.}(2020)]%
        {DBLP:conf/icml/ChenK0H20}
\bibfield{author}{\bibinfo{person}{Ting Chen}, \bibinfo{person}{Simon Kornblith}, \bibinfo{person}{Mohammad Norouzi}, {and} \bibinfo{person}{Geoffrey~E. Hinton}.} \bibinfo{year}{2020}\natexlab{}.
\newblock \showarticletitle{A Simple Framework for Contrastive Learning of Visual Representations}. In \bibinfo{booktitle}{\emph{ICML}}, Vol.~\bibinfo{volume}{119}. \bibinfo{pages}{1597--1607}.
\newblock


\bibitem[Dai et~al\mbox{.}(2017)]%
        {dai2017scannet}
\bibfield{author}{\bibinfo{person}{Angela Dai}, \bibinfo{person}{Angel~X Chang}, \bibinfo{person}{Manolis Savva}, \bibinfo{person}{Maciej Halber}, \bibinfo{person}{Thomas Funkhouser}, {and} \bibinfo{person}{Matthias Nie{\ss}ner}.} \bibinfo{year}{2017}\natexlab{}.
\newblock \showarticletitle{Scannet: Richly-annotated 3d reconstructions of indoor scenes}. In \bibinfo{booktitle}{\emph{Proceedings of the IEEE conference on computer vision and pattern recognition}}. \bibinfo{pages}{5828--5839}.
\newblock


\bibitem[Dong et~al\mbox{.}(2022)]%
        {dong2022autoencoders}
\bibfield{author}{\bibinfo{person}{Runpei Dong}, \bibinfo{person}{Zekun Qi}, \bibinfo{person}{Linfeng Zhang}, \bibinfo{person}{Junbo Zhang}, \bibinfo{person}{Jianjian Sun}, \bibinfo{person}{Zheng Ge}, \bibinfo{person}{Li Yi}, {and} \bibinfo{person}{Kaisheng Ma}.} \bibinfo{year}{2022}\natexlab{}.
\newblock \showarticletitle{Autoencoders as Cross-Modal Teachers: Can Pretrained {2D} Image Transformers Help {3D} Representation Learning?}
\newblock \bibinfo{journal}{\emph{CoRR}}  \bibinfo{volume}{abs/2212.08320} (\bibinfo{year}{2022}).
\newblock


\bibitem[Dosovitskiy et~al\mbox{.}(2021)]%
        {dosovitskiy2020image}
\bibfield{author}{\bibinfo{person}{Alexey Dosovitskiy}, \bibinfo{person}{Lucas Beyer}, \bibinfo{person}{Alexander Kolesnikov}, \bibinfo{person}{Dirk Weissenborn}, \bibinfo{person}{Xiaohua Zhai}, \bibinfo{person}{Thomas Unterthiner}, \bibinfo{person}{Mostafa Dehghani}, \bibinfo{person}{Matthias Minderer}, \bibinfo{person}{Georg Heigold}, \bibinfo{person}{Sylvain Gelly}, \bibinfo{person}{Jakob Uszkoreit}, {and} \bibinfo{person}{Neil Houlsby}.} \bibinfo{year}{2021}\natexlab{}.
\newblock \showarticletitle{An Image is Worth 16x16 Words: Transformers for Image Recognition at Scale}. In \bibinfo{booktitle}{\emph{ICLR}}.
\newblock


\bibitem[Fu et~al\mbox{.}(2023)]%
        {fu2023boosting}
\bibfield{author}{\bibinfo{person}{Kexue Fu}, \bibinfo{person}{Mingzhi Yuan}, \bibinfo{person}{Shaolei Liu}, {and} \bibinfo{person}{Manning Wang}.} \bibinfo{year}{2023}\natexlab{}.
\newblock \showarticletitle{Boosting Point-BERT by Multi-choice Tokens}.
\newblock \bibinfo{journal}{\emph{IEEE Transactions on Circuits and Systems for Video Technology}} (\bibinfo{year}{2023}).
\newblock


\bibitem[Geiger et~al\mbox{.}(2012)]%
        {geiger2012we}
\bibfield{author}{\bibinfo{person}{Andreas Geiger}, \bibinfo{person}{Philip Lenz}, {and} \bibinfo{person}{Raquel Urtasun}.} \bibinfo{year}{2012}\natexlab{}.
\newblock \showarticletitle{Are we ready for autonomous driving? the kitti vision benchmark suite}. In \bibinfo{booktitle}{\emph{2012 IEEE conference on computer vision and pattern recognition}}. IEEE, \bibinfo{pages}{3354--3361}.
\newblock


\bibitem[Guo et~al\mbox{.}(2021)]%
        {guo2021pct}
\bibfield{author}{\bibinfo{person}{Meng-Hao Guo}, \bibinfo{person}{Jun-Xiong Cai}, \bibinfo{person}{Zheng-Ning Liu}, \bibinfo{person}{Tai-Jiang Mu}, \bibinfo{person}{Ralph~R Martin}, {and} \bibinfo{person}{Shi-Min Hu}.} \bibinfo{year}{2021}\natexlab{}.
\newblock \showarticletitle{{PCT}: Point cloud transformer}.
\newblock \bibinfo{journal}{\emph{Computational Visual Media}} \bibinfo{volume}{7}, \bibinfo{number}{2} (\bibinfo{year}{2021}), \bibinfo{pages}{187--199}.
\newblock


\bibitem[Guo et~al\mbox{.}(2023)]%
        {DBLP:journals/corr/abs-2302-14007}
\bibfield{author}{\bibinfo{person}{Ziyu Guo}, \bibinfo{person}{Xianzhi Li}, {and} \bibinfo{person}{Pheng{-}Ann Heng}.} \bibinfo{year}{2023}\natexlab{}.
\newblock \showarticletitle{{Joint-MAE}: {2D-3D} Joint Masked Autoencoders for {3D} Point Cloud Pre-training}.
\newblock \bibinfo{journal}{\emph{CoRR}}  \bibinfo{volume}{abs/2302.14007} (\bibinfo{year}{2023}).
\newblock


\bibitem[Han et~al\mbox{.}(2019a)]%
        {han2019view}
\bibfield{author}{\bibinfo{person}{Zhizhong Han}, \bibinfo{person}{Mingyang Shang}, \bibinfo{person}{Yu-Shen Liu}, {and} \bibinfo{person}{Matthias Zwicker}.} \bibinfo{year}{2019}\natexlab{a}.
\newblock \showarticletitle{View inter-prediction {GAN}: Unsupervised representation learning for {3D} shapes by learning global shape memories to support local view predictions}. In \bibinfo{booktitle}{\emph{AAAI}}, Vol.~\bibinfo{volume}{33}. \bibinfo{pages}{8376--8384}.
\newblock


\bibitem[Han et~al\mbox{.}(2019b)]%
        {DBLP:conf/iccv/HanWLZ19}
\bibfield{author}{\bibinfo{person}{Zhizhong Han}, \bibinfo{person}{Xiyang Wang}, \bibinfo{person}{Yu{-}Shen Liu}, {and} \bibinfo{person}{Matthias Zwicker}.} \bibinfo{year}{2019}\natexlab{b}.
\newblock \showarticletitle{Multi-Angle Point Cloud-{VAE}: Unsupervised Feature Learning for {3D} Point Clouds From Multiple Angles by Joint Self-Reconstruction and Half-to-Half Prediction}. In \bibinfo{booktitle}{\emph{ICCV}}. \bibinfo{pages}{10441--10450}.
\newblock


\bibitem[Hassani and Haley(2019)]%
        {DBLP:conf/iccv/HassaniH19}
\bibfield{author}{\bibinfo{person}{Kaveh Hassani} {and} \bibinfo{person}{Mike Haley}.} \bibinfo{year}{2019}\natexlab{}.
\newblock \showarticletitle{Unsupervised Multi-Task Feature Learning on Point Clouds}. In \bibinfo{booktitle}{\emph{ICCV}}. \bibinfo{pages}{8159--8170}.
\newblock


\bibitem[He et~al\mbox{.}(2022)]%
        {he2022masked}
\bibfield{author}{\bibinfo{person}{Kaiming He}, \bibinfo{person}{Xinlei Chen}, \bibinfo{person}{Saining Xie}, \bibinfo{person}{Yanghao Li}, \bibinfo{person}{Piotr Doll{\'a}r}, {and} \bibinfo{person}{Ross Girshick}.} \bibinfo{year}{2022}\natexlab{}.
\newblock \showarticletitle{Masked autoencoders are scalable vision learners}. In \bibinfo{booktitle}{\emph{Proceedings of the IEEE/CVF Conference on Computer Vision and Pattern Recognition}}. \bibinfo{pages}{16000--16009}.
\newblock


\bibitem[He et~al\mbox{.}(2020)]%
        {KaimingHe2019MomentumCF}
\bibfield{author}{\bibinfo{person}{Kaiming He}, \bibinfo{person}{Haoqi Fan}, \bibinfo{person}{Yuxin Wu}, \bibinfo{person}{Saining Xie}, {and} \bibinfo{person}{Ross~B. Girshick}.} \bibinfo{year}{2020}\natexlab{}.
\newblock \showarticletitle{Momentum Contrast for Unsupervised Visual Representation Learning}. In \bibinfo{booktitle}{\emph{CVPR}}. \bibinfo{pages}{9726--9735}.
\newblock


\bibitem[Hou et~al\mbox{.}(2019)]%
        {hou20193d}
\bibfield{author}{\bibinfo{person}{Ji Hou}, \bibinfo{person}{Angela Dai}, {and} \bibinfo{person}{Matthias Nie{\ss}ner}.} \bibinfo{year}{2019}\natexlab{}.
\newblock \showarticletitle{3d-sis: 3d semantic instance segmentation of rgb-d scans}. In \bibinfo{booktitle}{\emph{Proceedings of the IEEE/CVF conference on computer vision and pattern recognition}}. \bibinfo{pages}{4421--4430}.
\newblock


\bibitem[Huang et~al\mbox{.}(2022)]%
        {huang2022clip2point}
\bibfield{author}{\bibinfo{person}{Tianyu Huang}, \bibinfo{person}{Bowen Dong}, \bibinfo{person}{Yunhan Yang}, \bibinfo{person}{Xiaoshui Huang}, \bibinfo{person}{Rynson W.~H. Lau}, \bibinfo{person}{Wanli Ouyang}, {and} \bibinfo{person}{Wangmeng Zuo}.} \bibinfo{year}{2022}\natexlab{}.
\newblock \showarticletitle{{CLIP2Point}: Transfer {CLIP} to Point Cloud Classification with Image-Depth Pre-training}.
\newblock \bibinfo{journal}{\emph{CoRR}}  \bibinfo{volume}{abs/2210.01055} (\bibinfo{year}{2022}).
\newblock


\bibitem[Kim et~al\mbox{.}(2021)]%
        {kim2021vilt}
\bibfield{author}{\bibinfo{person}{Wonjae Kim}, \bibinfo{person}{Bokyung Son}, {and} \bibinfo{person}{Ildoo Kim}.} \bibinfo{year}{2021}\natexlab{}.
\newblock \showarticletitle{{Vilt}: Vision-and-language transformer without convolution or region supervision}. In \bibinfo{booktitle}{\emph{ICML}}. \bibinfo{pages}{5583--5594}.
\newblock


\bibitem[Kingma and Ba(2015)]%
        {DiederikPKingma2014AdamAM}
\bibfield{author}{\bibinfo{person}{Diederik~P. Kingma} {and} \bibinfo{person}{Jimmy Ba}.} \bibinfo{year}{2015}\natexlab{}.
\newblock \showarticletitle{{Adam}: {A} Method for Stochastic Optimization}. In \bibinfo{booktitle}{\emph{ICLR}}.
\newblock


\bibitem[Li et~al\mbox{.}(2021)]%
        {li2021align}
\bibfield{author}{\bibinfo{person}{Junnan Li}, \bibinfo{person}{Ramprasaath Selvaraju}, \bibinfo{person}{Akhilesh Gotmare}, \bibinfo{person}{Shafiq Joty}, \bibinfo{person}{Caiming Xiong}, {and} \bibinfo{person}{Steven Chu~Hong Hoi}.} \bibinfo{year}{2021}\natexlab{}.
\newblock \showarticletitle{Align before fuse: Vision and language representation learning with momentum distillation}.
\newblock \bibinfo{journal}{\emph{NeurIPS}}  \bibinfo{volume}{34} (\bibinfo{year}{2021}), \bibinfo{pages}{9694--9705}.
\newblock


\bibitem[Li et~al\mbox{.}(2022)]%
        {li2022mc}
\bibfield{author}{\bibinfo{person}{Xiaotong Li}, \bibinfo{person}{Yixiao Ge}, \bibinfo{person}{Kun Yi}, \bibinfo{person}{Zixuan Hu}, \bibinfo{person}{Ying Shan}, {and} \bibinfo{person}{Ling-Yu Duan}.} \bibinfo{year}{2022}\natexlab{}.
\newblock \showarticletitle{{mc-BEiT}: Multi-choice discretization for image bert pre-training}. In \bibinfo{booktitle}{\emph{ECCV}}. \bibinfo{pages}{231--246}.
\newblock


\bibitem[Li et~al\mbox{.}(2018)]%
        {DBLP:conf/nips/LiBSWDC18}
\bibfield{author}{\bibinfo{person}{Yangyan Li}, \bibinfo{person}{Rui Bu}, \bibinfo{person}{Mingchao Sun}, \bibinfo{person}{Wei Wu}, \bibinfo{person}{Xinhan Di}, {and} \bibinfo{person}{Baoquan Chen}.} \bibinfo{year}{2018}\natexlab{}.
\newblock \showarticletitle{{PointCNN}: Convolution On X-Transformed Points}. In \bibinfo{booktitle}{\emph{NeurIPS}}. \bibinfo{pages}{828--838}.
\newblock


\bibitem[Liu et~al\mbox{.}(2022)]%
        {liu2022masked}
\bibfield{author}{\bibinfo{person}{Haotian Liu}, \bibinfo{person}{Mu Cai}, {and} \bibinfo{person}{Yong~Jae Lee}.} \bibinfo{year}{2022}\natexlab{}.
\newblock \showarticletitle{Masked discrimination for self-supervised learning on point clouds}. In \bibinfo{booktitle}{\emph{ECCV}}. \bibinfo{pages}{657--675}.
\newblock


\bibitem[Liu et~al\mbox{.}(2023)]%
        {10314552}
\bibfield{author}{\bibinfo{person}{Yun Liu}, \bibinfo{person}{Xuefeng Yan}, \bibinfo{person}{Zhiqi Li}, \bibinfo{person}{Zhilei Chen}, \bibinfo{person}{Zeyong Wei}, {and} \bibinfo{person}{Mingqiang Wei}.} \bibinfo{year}{2023}\natexlab{}.
\newblock \showarticletitle{PointGame: Geometrically and Adaptively Masked Auto-Encoder on Point Clouds}.
\newblock \bibinfo{journal}{\emph{IEEE Transactions on Geoscience and Remote Sensing}} (\bibinfo{year}{2023}), \bibinfo{pages}{1--1}.
\newblock
\urldef\tempurl%
\url{https://doi.org/10.1109/TGRS.2023.3331748}
\showDOI{\tempurl}


\bibitem[Lu et~al\mbox{.}(2019)]%
        {lu2019vilbert}
\bibfield{author}{\bibinfo{person}{Jiasen Lu}, \bibinfo{person}{Dhruv Batra}, \bibinfo{person}{Devi Parikh}, {and} \bibinfo{person}{Stefan Lee}.} \bibinfo{year}{2019}\natexlab{}.
\newblock \showarticletitle{{Vilbert}: Pretraining task-agnostic visiolinguistic representations for vision-and-language tasks}.
\newblock \bibinfo{journal}{\emph{NeurIPS}}  \bibinfo{volume}{32} (\bibinfo{year}{2019}).
\newblock


\bibitem[Ma et~al\mbox{.}(2022)]%
        {ma2022rethinking}
\bibfield{author}{\bibinfo{person}{Xu Ma}, \bibinfo{person}{Can Qin}, \bibinfo{person}{Haoxuan You}, \bibinfo{person}{Haoxi Ran}, {and} \bibinfo{person}{Yun Fu}.} \bibinfo{year}{2022}\natexlab{}.
\newblock \showarticletitle{Rethinking Network Design and Local Geometry in Point Cloud: {A} Simple Residual {MLP} Framework}. In \bibinfo{booktitle}{\emph{ICLR}}.
\newblock


\bibitem[Misra et~al\mbox{.}(2021)]%
        {demisra2021end}
\bibfield{author}{\bibinfo{person}{Ishan Misra}, \bibinfo{person}{Rohit Girdhar}, {and} \bibinfo{person}{Armand Joulin}.} \bibinfo{year}{2021}\natexlab{}.
\newblock \showarticletitle{An end-to-end transformer model for 3d object detection}. In \bibinfo{booktitle}{\emph{Proceedings of the IEEE/CVF International Conference on Computer Vision}}. \bibinfo{pages}{2906--2917}.
\newblock


\bibitem[Pang et~al\mbox{.}(2022)]%
        {pang2022masked}
\bibfield{author}{\bibinfo{person}{Yatian Pang}, \bibinfo{person}{Wenxiao Wang}, \bibinfo{person}{Francis~EH Tay}, \bibinfo{person}{Wei Liu}, \bibinfo{person}{Yonghong Tian}, {and} \bibinfo{person}{Li Yuan}.} \bibinfo{year}{2022}\natexlab{}.
\newblock \showarticletitle{Masked autoencoders for point cloud self-supervised learning}. In \bibinfo{booktitle}{\emph{ECCV}}. \bibinfo{pages}{604--621}.
\newblock


\bibitem[Peng et~al\mbox{.}(2022)]%
        {peng2022beit}
\bibfield{author}{\bibinfo{person}{Zhiliang Peng}, \bibinfo{person}{Li Dong}, \bibinfo{person}{Hangbo Bao}, \bibinfo{person}{Qixiang Ye}, {and} \bibinfo{person}{Furu Wei}.} \bibinfo{year}{2022}\natexlab{}.
\newblock \showarticletitle{{BEiT} v2: Masked Image Modeling with Vector-Quantized Visual Tokenizers}.
\newblock \bibinfo{journal}{\emph{CoRR}}  \bibinfo{volume}{abs/2208.06366} (\bibinfo{year}{2022}).
\newblock


\bibitem[Qi et~al\mbox{.}(2019)]%
        {qi2019deep}
\bibfield{author}{\bibinfo{person}{Charles~R Qi}, \bibinfo{person}{Or Litany}, \bibinfo{person}{Kaiming He}, {and} \bibinfo{person}{Leonidas~J Guibas}.} \bibinfo{year}{2019}\natexlab{}.
\newblock \showarticletitle{Deep hough voting for 3d object detection in point clouds}. In \bibinfo{booktitle}{\emph{proceedings of the IEEE/CVF International Conference on Computer Vision}}. \bibinfo{pages}{9277--9286}.
\newblock


\bibitem[Qi et~al\mbox{.}(2017a)]%
        {DBLP:conf/cvpr/QiSMG17}
\bibfield{author}{\bibinfo{person}{Charles~Ruizhongtai Qi}, \bibinfo{person}{Hao Su}, \bibinfo{person}{Kaichun Mo}, {and} \bibinfo{person}{Leonidas~J. Guibas}.} \bibinfo{year}{2017}\natexlab{a}.
\newblock \showarticletitle{{PointNet}: Deep Learning on Point Sets for {3D} Classification and Segmentation}. In \bibinfo{booktitle}{\emph{CVPR}}. \bibinfo{pages}{77--85}.
\newblock


\bibitem[Qi et~al\mbox{.}(2017b)]%
        {DBLP:conf/nips/QiYSG17}
\bibfield{author}{\bibinfo{person}{Charles~Ruizhongtai Qi}, \bibinfo{person}{Li Yi}, \bibinfo{person}{Hao Su}, {and} \bibinfo{person}{Leonidas~J. Guibas}.} \bibinfo{year}{2017}\natexlab{b}.
\newblock \showarticletitle{{PointNet++}: Deep Hierarchical Feature Learning on Point Sets in a Metric Space}. In \bibinfo{booktitle}{\emph{NeurIPS}}. \bibinfo{pages}{5099--5108}.
\newblock


\bibitem[Radford et~al\mbox{.}(2021)]%
        {radford2021learning}
\bibfield{author}{\bibinfo{person}{Alec Radford}, \bibinfo{person}{Jong~Wook Kim}, \bibinfo{person}{Chris Hallacy}, \bibinfo{person}{Aditya Ramesh}, \bibinfo{person}{Gabriel Goh}, \bibinfo{person}{Sandhini Agarwal}, \bibinfo{person}{Girish Sastry}, \bibinfo{person}{Amanda Askell}, \bibinfo{person}{Pamela Mishkin}, \bibinfo{person}{Jack Clark}, {et~al\mbox{.}}} \bibinfo{year}{2021}\natexlab{}.
\newblock \showarticletitle{Learning transferable visual models from natural language supervision}. In \bibinfo{booktitle}{\emph{ICML}}. \bibinfo{pages}{8748--8763}.
\newblock


\bibitem[Ramesh et~al\mbox{.}(2021)]%
        {ramesh2021zero}
\bibfield{author}{\bibinfo{person}{Aditya Ramesh}, \bibinfo{person}{Mikhail Pavlov}, \bibinfo{person}{Gabriel Goh}, \bibinfo{person}{Scott Gray}, \bibinfo{person}{Chelsea Voss}, \bibinfo{person}{Alec Radford}, \bibinfo{person}{Mark Chen}, {and} \bibinfo{person}{Ilya Sutskever}.} \bibinfo{year}{2021}\natexlab{}.
\newblock \showarticletitle{Zero-shot text-to-image generation}. In \bibinfo{booktitle}{\emph{ICML}}. \bibinfo{pages}{8821--8831}.
\newblock


\bibitem[Sharma and Kaul(2020)]%
        {sharma2020self}
\bibfield{author}{\bibinfo{person}{Charu Sharma} {and} \bibinfo{person}{Manohar Kaul}.} \bibinfo{year}{2020}\natexlab{}.
\newblock \showarticletitle{Self-supervised few-shot learning on point clouds}.
\newblock \bibinfo{journal}{\emph{NeurIPS}}  \bibinfo{volume}{33} (\bibinfo{year}{2020}), \bibinfo{pages}{7212--7221}.
\newblock


\bibitem[Song et~al\mbox{.}(2015)]%
        {song2015sun}
\bibfield{author}{\bibinfo{person}{Shuran Song}, \bibinfo{person}{Samuel~P Lichtenberg}, {and} \bibinfo{person}{Jianxiong Xiao}.} \bibinfo{year}{2015}\natexlab{}.
\newblock \showarticletitle{Sun rgb-d: A rgb-d scene understanding benchmark suite}. In \bibinfo{booktitle}{\emph{Proceedings of the IEEE conference on computer vision and pattern recognition}}. \bibinfo{pages}{567--576}.
\newblock


\bibitem[Tian et~al\mbox{.}(2020)]%
        {tian2020rethinking}
\bibfield{author}{\bibinfo{person}{Yonglong Tian}, \bibinfo{person}{Yue Wang}, \bibinfo{person}{Dilip Krishnan}, \bibinfo{person}{Joshua~B Tenenbaum}, {and} \bibinfo{person}{Phillip Isola}.} \bibinfo{year}{2020}\natexlab{}.
\newblock \showarticletitle{Rethinking few-shot image classification: a good embedding is all you need?}. In \bibinfo{booktitle}{\emph{Computer Vision--ECCV 2020: 16th European Conference, Glasgow, UK, August 23--28, 2020, Proceedings, Part XIV 16}}. Springer, \bibinfo{pages}{266--282}.
\newblock


\bibitem[Uy et~al\mbox{.}(2019)]%
        {DBLP:conf/iccv/UyPHNY19}
\bibfield{author}{\bibinfo{person}{Mikaela~Angelina Uy}, \bibinfo{person}{Quang{-}Hieu Pham}, \bibinfo{person}{Binh{-}Son Hua}, \bibinfo{person}{Duc~Thanh Nguyen}, {and} \bibinfo{person}{Sai{-}Kit Yeung}.} \bibinfo{year}{2019}\natexlab{}.
\newblock \showarticletitle{Revisiting Point Cloud Classification: {A} New Benchmark Dataset and Classification Model on Real-World Data}. In \bibinfo{booktitle}{\emph{ICCV}}. \bibinfo{pages}{1588--1597}.
\newblock


\bibitem[Vaswani et~al\mbox{.}(2017)]%
        {vaswani2017attention}
\bibfield{author}{\bibinfo{person}{Ashish Vaswani}, \bibinfo{person}{Noam Shazeer}, \bibinfo{person}{Niki Parmar}, \bibinfo{person}{Jakob Uszkoreit}, \bibinfo{person}{Llion Jones}, \bibinfo{person}{Aidan~N Gomez}, \bibinfo{person}{{\L}ukasz Kaiser}, {and} \bibinfo{person}{Illia Polosukhin}.} \bibinfo{year}{2017}\natexlab{}.
\newblock \showarticletitle{Attention is all you need}.
\newblock \bibinfo{journal}{\emph{NeurIPS}}  \bibinfo{volume}{30} (\bibinfo{year}{2017}).
\newblock


\bibitem[Wang et~al\mbox{.}(2021)]%
        {wang2021unsupervised}
\bibfield{author}{\bibinfo{person}{Hanchen Wang}, \bibinfo{person}{Qi Liu}, \bibinfo{person}{Xiangyu Yue}, \bibinfo{person}{Joan Lasenby}, {and} \bibinfo{person}{Matt~J Kusner}.} \bibinfo{year}{2021}\natexlab{}.
\newblock \showarticletitle{Unsupervised point cloud pre-training via occlusion completion}. In \bibinfo{booktitle}{\emph{ICCV}}. \bibinfo{pages}{9782--9792}.
\newblock


\bibitem[Wang et~al\mbox{.}(2019)]%
        {DBLP:journals/tog/WangSLSBS19}
\bibfield{author}{\bibinfo{person}{Yue Wang}, \bibinfo{person}{Yongbin Sun}, \bibinfo{person}{Ziwei Liu}, \bibinfo{person}{Sanjay~E. Sarma}, \bibinfo{person}{Michael~M. Bronstein}, {and} \bibinfo{person}{Justin~M. Solomon}.} \bibinfo{year}{2019}\natexlab{}.
\newblock \showarticletitle{Dynamic Graph {CNN} for Learning on Point Clouds}.
\newblock \bibinfo{journal}{\emph{{ACM} Trans. Graph.}} \bibinfo{volume}{38}, \bibinfo{number}{5} (\bibinfo{year}{2019}), \bibinfo{pages}{146:1--146:12}.
\newblock


\bibitem[Wang et~al\mbox{.}(2022)]%
        {wang2022p2p}
\bibfield{author}{\bibinfo{person}{Ziyi Wang}, \bibinfo{person}{Xumin Yu}, \bibinfo{person}{Yongming Rao}, \bibinfo{person}{Jie Zhou}, {and} \bibinfo{person}{Jiwen Lu}.} \bibinfo{year}{2022}\natexlab{}.
\newblock \showarticletitle{{P2P:} Tuning Pre-trained Image Models for Point Cloud Analysis with Point-to-Pixel Prompting}. In \bibinfo{booktitle}{\emph{NeurIPS}}.
\newblock


\bibitem[Wei et~al\mbox{.}(2022)]%
        {wei2022masked}
\bibfield{author}{\bibinfo{person}{Chen Wei}, \bibinfo{person}{Haoqi Fan}, \bibinfo{person}{Saining Xie}, \bibinfo{person}{Chao-Yuan Wu}, \bibinfo{person}{Alan Yuille}, {and} \bibinfo{person}{Christoph Feichtenhofer}.} \bibinfo{year}{2022}\natexlab{}.
\newblock \showarticletitle{Masked feature prediction for self-supervised visual pre-training}. In \bibinfo{booktitle}{\emph{Proceedings of the IEEE/CVF Conference on Computer Vision and Pattern Recognition}}. \bibinfo{pages}{14668--14678}.
\newblock


\bibitem[Wu et~al\mbox{.}(2015)]%
        {DBLP:conf/cvpr/WuSKYZTX15}
\bibfield{author}{\bibinfo{person}{Zhirong Wu}, \bibinfo{person}{Shuran Song}, \bibinfo{person}{Aditya Khosla}, \bibinfo{person}{Fisher Yu}, \bibinfo{person}{Linguang Zhang}, \bibinfo{person}{Xiaoou Tang}, {and} \bibinfo{person}{Jianxiong Xiao}.} \bibinfo{year}{2015}\natexlab{}.
\newblock \showarticletitle{{3D ShapeNets}: {A} deep representation for volumetric shapes}. In \bibinfo{booktitle}{\emph{CVPR}}. \bibinfo{pages}{1912--1920}.
\newblock


\bibitem[Xie et~al\mbox{.}(2020)]%
        {xie2020pointcontrast}
\bibfield{author}{\bibinfo{person}{Saining Xie}, \bibinfo{person}{Jiatao Gu}, \bibinfo{person}{Demi Guo}, \bibinfo{person}{Charles~R Qi}, \bibinfo{person}{Leonidas Guibas}, {and} \bibinfo{person}{Or Litany}.} \bibinfo{year}{2020}\natexlab{}.
\newblock \showarticletitle{{PointContrast}: Unsupervised pre-training for {3D} point cloud understanding}. In \bibinfo{booktitle}{\emph{ECCV}}. \bibinfo{pages}{574--591}.
\newblock


\bibitem[Xie et~al\mbox{.}(2018)]%
        {DBLP:conf/cvpr/XieLCT18}
\bibfield{author}{\bibinfo{person}{Saining Xie}, \bibinfo{person}{Sainan Liu}, \bibinfo{person}{Zeyu Chen}, {and} \bibinfo{person}{Zhuowen Tu}.} \bibinfo{year}{2018}\natexlab{}.
\newblock \showarticletitle{Attentional ShapeContextNet for Point Cloud Recognition}. In \bibinfo{booktitle}{\emph{CVPR}}. \bibinfo{pages}{4606--4615}.
\newblock


\bibitem[Xu et~al\mbox{.}(2018)]%
        {xu2018pointfusion}
\bibfield{author}{\bibinfo{person}{Danfei Xu}, \bibinfo{person}{Dragomir Anguelov}, {and} \bibinfo{person}{Ashesh Jain}.} \bibinfo{year}{2018}\natexlab{}.
\newblock \showarticletitle{Pointfusion: Deep sensor fusion for 3d bounding box estimation}. In \bibinfo{booktitle}{\emph{Proceedings of the IEEE conference on computer vision and pattern recognition}}. \bibinfo{pages}{244--253}.
\newblock


\bibitem[Yang et~al\mbox{.}(2022)]%
        {yang2022vision}
\bibfield{author}{\bibinfo{person}{Jinyu Yang}, \bibinfo{person}{Jiali Duan}, \bibinfo{person}{Son Tran}, \bibinfo{person}{Yi Xu}, \bibinfo{person}{Sampath Chanda}, \bibinfo{person}{Liqun Chen}, \bibinfo{person}{Belinda Zeng}, \bibinfo{person}{Trishul Chilimbi}, {and} \bibinfo{person}{Junzhou Huang}.} \bibinfo{year}{2022}\natexlab{}.
\newblock \showarticletitle{Vision-language pre-training with triple contrastive learning}. In \bibinfo{booktitle}{\emph{CVPR}}. \bibinfo{pages}{15671--15680}.
\newblock


\bibitem[Yang et~al\mbox{.}(2018)]%
        {DBLP:conf/cvpr/YangFST18}
\bibfield{author}{\bibinfo{person}{Yaoqing Yang}, \bibinfo{person}{Chen Feng}, \bibinfo{person}{Yiru Shen}, {and} \bibinfo{person}{Dong Tian}.} \bibinfo{year}{2018}\natexlab{}.
\newblock \showarticletitle{{FoldingNet}: Point Cloud Auto-Encoder via Deep Grid Deformation}. In \bibinfo{booktitle}{\emph{CVPR}}. \bibinfo{pages}{206--215}.
\newblock


\bibitem[Yao et~al\mbox{.}(2022)]%
        {yao2021filip}
\bibfield{author}{\bibinfo{person}{Lewei Yao}, \bibinfo{person}{Runhui Huang}, \bibinfo{person}{Lu Hou}, \bibinfo{person}{Guansong Lu}, \bibinfo{person}{Minzhe Niu}, \bibinfo{person}{Hang Xu}, \bibinfo{person}{Xiaodan Liang}, \bibinfo{person}{Zhenguo Li}, \bibinfo{person}{Xin Jiang}, {and} \bibinfo{person}{Chunjing Xu}.} \bibinfo{year}{2022}\natexlab{}.
\newblock \showarticletitle{{FILIP:} Fine-grained Interactive Language-Image Pre-Training}. In \bibinfo{booktitle}{\emph{ICLR}}.
\newblock


\bibitem[Yi et~al\mbox{.}(2016)]%
        {DBLP:journals/tog/YiKCSYSLHSG16}
\bibfield{author}{\bibinfo{person}{Li Yi}, \bibinfo{person}{Vladimir~G. Kim}, \bibinfo{person}{Duygu Ceylan}, \bibinfo{person}{I{-}Chao Shen}, \bibinfo{person}{Mengyan Yan}, \bibinfo{person}{Hao Su}, \bibinfo{person}{Cewu Lu}, \bibinfo{person}{Qixing Huang}, \bibinfo{person}{Alla Sheffer}, {and} \bibinfo{person}{Leonidas~J. Guibas}.} \bibinfo{year}{2016}\natexlab{}.
\newblock \showarticletitle{A scalable active framework for region annotation in {3D} shape collections}.
\newblock \bibinfo{journal}{\emph{{ACM} Trans. Graph.}} \bibinfo{volume}{35}, \bibinfo{number}{6} (\bibinfo{year}{2016}), \bibinfo{pages}{210:1--210:12}.
\newblock


\bibitem[Yu et~al\mbox{.}(2022)]%
        {DBLP:conf/cvpr/YuTR00L22}
\bibfield{author}{\bibinfo{person}{Xumin Yu}, \bibinfo{person}{Lulu Tang}, \bibinfo{person}{Yongming Rao}, \bibinfo{person}{Tiejun Huang}, \bibinfo{person}{Jie Zhou}, {and} \bibinfo{person}{Jiwen Lu}.} \bibinfo{year}{2022}\natexlab{}.
\newblock \showarticletitle{{Point-BERT}: Pre-training {3D} Point Cloud Transformers with Masked Point Modeling}. In \bibinfo{booktitle}{\emph{CVPR}}. \bibinfo{pages}{19291--19300}.
\newblock


\bibitem[Zhang and Zhu(2019)]%
        {DBLP:conf/3dim/ZhangZ19}
\bibfield{author}{\bibinfo{person}{Ling Zhang} {and} \bibinfo{person}{Zhigang Zhu}.} \bibinfo{year}{2019}\natexlab{}.
\newblock \showarticletitle{Unsupervised Feature Learning for Point Cloud Understanding by Contrasting and Clustering Using Graph Convolutional Neural Networks}. In \bibinfo{booktitle}{\emph{3DV}}. \bibinfo{pages}{395--404}.
\newblock


\bibitem[Zhang et~al\mbox{.}(2022a)]%
        {zhang2022point}
\bibfield{author}{\bibinfo{person}{Renrui Zhang}, \bibinfo{person}{Ziyu Guo}, \bibinfo{person}{Peng Gao}, \bibinfo{person}{Rongyao Fang}, \bibinfo{person}{Bin Zhao}, \bibinfo{person}{Dong Wang}, \bibinfo{person}{Yu Qiao}, {and} \bibinfo{person}{Hongsheng Li}.} \bibinfo{year}{2022}\natexlab{a}.
\newblock \showarticletitle{{Point-M2AE}: Multi-scale Masked Autoencoders for Hierarchical Point Cloud Pre-training}. In \bibinfo{booktitle}{\emph{NeurIPS}}.
\newblock


\bibitem[Zhang et~al\mbox{.}(2022b)]%
        {zhang2022monodetr}
\bibfield{author}{\bibinfo{person}{Renrui Zhang}, \bibinfo{person}{Han Qiu}, \bibinfo{person}{Tai Wang}, \bibinfo{person}{Xuanzhuo Xu}, \bibinfo{person}{Ziyu Guo}, \bibinfo{person}{Yu Qiao}, \bibinfo{person}{Peng Gao}, {and} \bibinfo{person}{Hongsheng Li}.} \bibinfo{year}{2022}\natexlab{b}.
\newblock \showarticletitle{Monodetr: Depth-aware transformer for monocular 3d object detection}.
\newblock \bibinfo{journal}{\emph{arXiv preprint arXiv:2203.13310}} \bibinfo{volume}{2}, \bibinfo{number}{3} (\bibinfo{year}{2022}), \bibinfo{pages}{7}.
\newblock


\bibitem[Zhao et~al\mbox{.}(2021)]%
        {DBLP:conf/iccv/ZhaoJJTK21}
\bibfield{author}{\bibinfo{person}{Hengshuang Zhao}, \bibinfo{person}{Li Jiang}, \bibinfo{person}{Jiaya Jia}, \bibinfo{person}{Philip H.~S. Torr}, {and} \bibinfo{person}{Vladlen Koltun}.} \bibinfo{year}{2021}\natexlab{}.
\newblock \showarticletitle{Point Transformer}. In \bibinfo{booktitle}{\emph{ICCV}}. \bibinfo{pages}{16239--16248}.
\newblock


\bibitem[Zhao et~al\mbox{.}(2019)]%
        {YonghengZhao20183DPC}
\bibfield{author}{\bibinfo{person}{Yongheng Zhao}, \bibinfo{person}{Tolga Birdal}, \bibinfo{person}{Haowen Deng}, {and} \bibinfo{person}{Federico Tombari}.} \bibinfo{year}{2019}\natexlab{}.
\newblock \showarticletitle{{3D} Point Capsule Networks}. In \bibinfo{booktitle}{\emph{CVPR}}. \bibinfo{pages}{1009--1018}.
\newblock


\end{thebibliography}


\end{document}